\documentclass[10pt]{article}

\usepackage{PRIMEarxiv}

\usepackage[utf8]{inputenc} 
\usepackage[T1]{fontenc}    
\usepackage{hyperref}       
\usepackage{url}            
\usepackage{booktabs}       
\usepackage{amsfonts}       
\usepackage{nicefrac}       
\usepackage{enumitem}
\usepackage{lipsum}
\usepackage{multirow}
\usepackage[linesnumbered,ruled,vlined]{algorithm2e}
\SetKwInput{Input}{Input}   
\SetKwInput{Output}{Output} 
\usepackage{amsmath, amssymb, mathrsfs, bbm}
\usepackage{tabularx}
\usepackage{setspace}
\usepackage{subcaption}
\usepackage{xcolor}         
\usepackage{fancyhdr}       
\usepackage{graphicx}       
\usepackage[backend=biber,style=ieee,sorting=none,natbib=true]{biblatex}
\usepackage{authblk}

\graphicspath{{media/}}     
\usepackage{bm}

\title{Real-Time Localization and Bimodal Point Pattern Analysis of Palms Using UAV Imagery
}

\author[1,2,*]{Kangning Cui}
\author[1,2,*]{Wei Tang}
\author[3]{Rongkun Zhu}
\author[4]{Manqi Wang}
\author[4]{Gregory D. Larsen}
\author[5]{Victor P. Pauca}
\author[5]{Sarra Alqahtani}
\author[5]{Fan Yang}
\author[4]{David Segurado}
\author[6]{Paul Fine}
\author[7]{Jordan Karubian}
\author[2,8,9]{Raymond H. Chan}
\author[5]{Robert J. Plemmons}
\author[1]{Jean-Michel Morel}
\author[4]{Miles R. Silman}

\affil[1]{Department of Mathematics, City University of Hong Kong}
\affil[2]{Hong Kong Centre for Cerebro-Cardiovascular Health Engineering, Hong Kong}
\affil[3]{Department of Computer Science, Xidian University, Xi'an, Shaanxi, China}
\affil[4]{Department of Biology, Wake Forest University, Winston-Salem, NC, USA}
\affil[5]{Department of Computer Science, Wake Forest University, Winston-Salem, NC, USA}
\affil[6]{Department of Integrative Biology, University of California, Berkeley, CA, USA}
\affil[7]{Department of Ecology and Evolutionary Biology, Tulane University, New Orleans, LA, USA}
\affil[8]{Department of Operations and Risk Management, Lingnan University, Tuen Mun, Hong Kong}
\affil[9]{School of Data Science, Lingnan University, Tuen Mun, Hong Kong}

\begin{document}
\maketitle

\let\thefootnote\relax
\noindent\footnotetext{$*$ Kangning Cui and Wei Tang contributed equally to this work. Corresponding author: kangnicui2-c@my.cityu.edu.hk}

\onehalfspacing
\begin{abstract}
Understanding the spatial distribution of palms within tropical forests is essential for effective ecological monitoring, conservation strategies, and the sustainable integration of natural forest products into local and global supply chains. However, the analysis of remotely sensed data in these environments faces significant challenges, such as overlapping palm and tree crowns, uneven shading across the canopy surface, and the heterogeneous nature of the forest landscapes, which often affect the performance of palm detection and segmentation algorithms. To overcome these issues, we introduce PalmDSNet, a deep learning framework for real-time detection, segmentation, and counting of canopy palms. Additionally, we employ a bimodal reproduction algorithm that simulates palm spatial propagation to further enhance the understanding of these point patterns using PalmDSNet's results. We used UAV-captured imagery to create orthomosaics from 21 sites across western Ecuadorian tropical forests, covering a gradient from the everwet Chocó forests near Colombia to the drier forests of southwestern Ecuador. Expert annotations were used to create a comprehensive dataset, including 7,356 bounding boxes on image patches and 7,603 palm centers across five orthomosaics, encompassing a total area of 449 hectares. By combining PalmDSNet with the bimodal reproduction algorithm, which optimizes parameters for both local and global spatial variability, we effectively simulate the spatial distribution of palms in diverse and dense tropical environments, validating its utility for advanced applications in tropical forest monitoring and remote sensing analysis. The dataset can be accessed at \href{https://zenodo.org/records/13822508}{10.5281/zenodo.13822508}, and the code to replicate the study is available at \href{https://github.com/ckn3/palm-ds-sp}{github.com/ckn3/palm-ds-sp}.

\end{abstract}


\section{Introduction}
\begin{figure}[tb]
  \centering
  \begin{subfigure}{0.49\linewidth}
    \includegraphics[width=\textwidth]{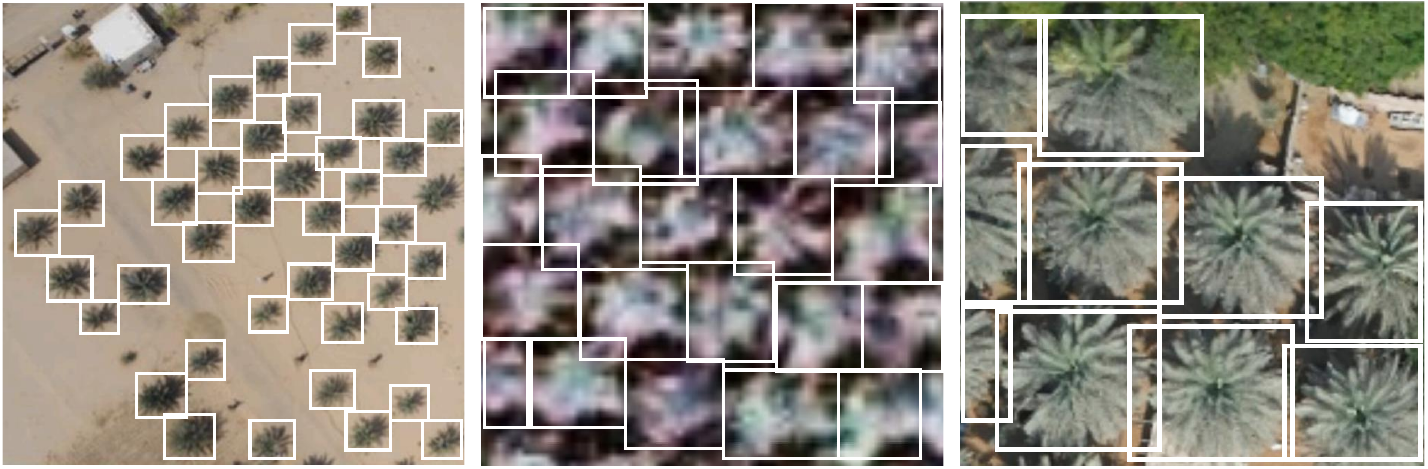}
    \caption{Cases from existing studies}
  \end{subfigure}%
  \begin{subfigure}{0.49\linewidth}
    \includegraphics[width=\textwidth]{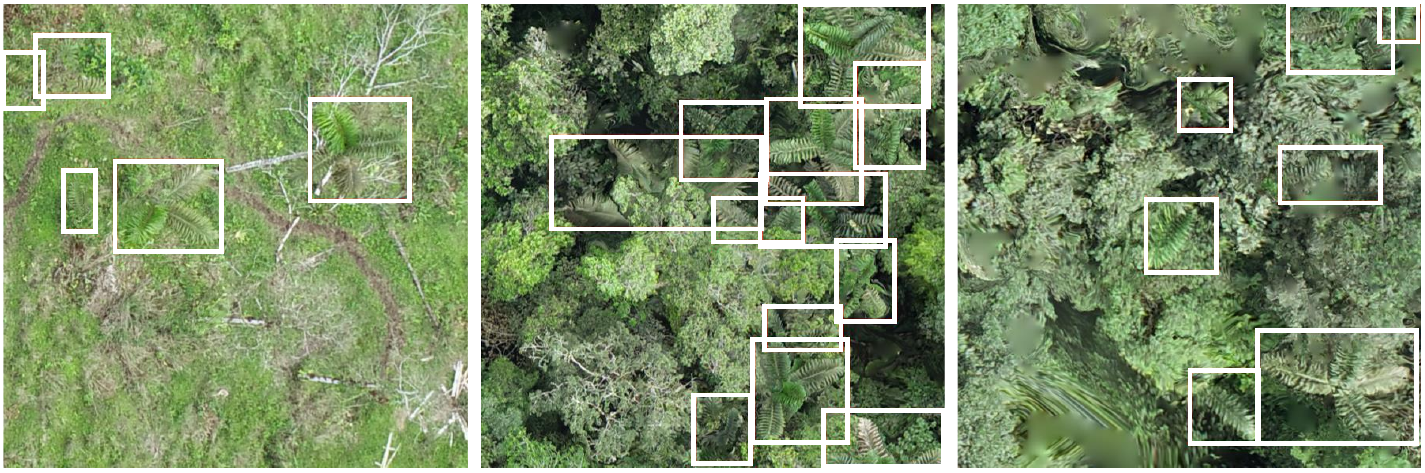}
    \caption{Cases from our dataset}
  \end{subfigure}
  \caption{Comparative Samples of Manual Labels.}
  \label{fig:manual}
\end{figure}

Palms (family \textit{Arecaceae}) include many ecologically and economically important species whose spatial distributions crucially inform tropical forest ecology and conservation research. They are also central to local economies and regional to global efforts to incorporate natural forest products into sustainable livelihoods and local to international forest product supply chains~\cite{eiserhardt2011geographical, zambrana2007diversity}. Tropical forests host a significant portion of global biodiversity and are increasingly threatened by deforestation and degradation~\cite{sutherland2013identification, camalan2022change}, and palms, with their distinctive ecological importance, can serve as vital indicators of both forest health and human impact, offering insights into biodiversity, soil quality, and the overall health of forest ecosystems~\cite{wagner2020regional}. They play a central role in shelter, food, and fiber, and are an emerging resource in the development of non-timber forest product markets, supporting human communities in indigenous and rural areas.  Palms also constitute essential and often keystone resources for tropical wildlife~\cite{pitman2014distribution, malhi2014tropical, van2019palm, terborgh1986community}. Here, we are concerned with identifying, locating, and quantifying palms occurring naturally within tropical forests, with particular attention to their spatial distribution and natural reproduction (see Figure~\ref{fig:manual}b), as opposed to palm plantations (see Figure~\ref{fig:manual}a). Knowledge of the spatial distribution and abundance of palms can inform sustainable use and management, leading to economic benefits to local communities. Thus, these tasks are crucial for effective management, economic development, and conservation, as well as for understanding basic ecological questions of palm distribution and abundance. 

Palms can be detected in high-resolution remotely sensed imagery by their distinctive leaves and crowns. While object detection and spatial pattern analysis are well-established in computer vision and statistics, applying these techniques to complex real-world environments -- such as identifying naturally occurring palms in tropical forests and analyzing their spatial distribution -- poses significant challenges~\cite{cui2024palmprobnet, tagle2019identifying}. First, palm species are found in highly imbalanced numbers, with certain species being far more prevalent than others~\cite{li2016deep, freudenberg2019large}. Although various species exist, typically only 2 or 3 species have sufficient number of samples  to enable reliable detection. Efficiently detecting all palm species, even the less common ones, is crucial for forest ecology.

Secondly, UAV optical imagery of tropical forests exhibits considerable variability in lighting, occlusion, and background clutter, complicating the analysis process~\cite{park2019quantifying, cui2024palmprobnet}. Palms are often obscured by neighboring tree canopies, making them difficult to detect. Moreover, the dense vegetation in the background and inconsistent lighting across different forest areas further complicates feature detection and localization. Shadows cast by dense canopies and differing sun angles throughout the day lead to significant changes in image brightness and contrast across sampled times and locations, challenging the consistency of algorithmic interpretation~\cite{gibril2021deep, cui2021unsupervised}.

Third, high-quality labeled datasets for realistic environments are notably scarce. The collection and detailed annotation of such data necessitate extensive fieldwork by skilled experts, which is particularly demanding in tropical forests like the Amazon~\cite{muscarella2020global, hidalgo2022sustainable, cui2024superpixel}. The process of annotating and labeling such data typically involves assembling raw imagery into orthomosaics, which are spatially referenced data products derived from UAV captures. However, this process is fraught with challenges. Orthomosaics often suffer from noise and artifacts due to image stitching (alignment, merging, and rectification) errors, varying lighting conditions, sensor discrepancies, and environmental factors such as wind and cloud cover, which can lead to movement and shading between images~\cite{zhang2023aerial}.

Moreover, real-time object detection methods for UAV-based remote sensing images are often underutilized, especially in large, densely forested regions where substantial computational resources may be lacking~\cite{qin2021ag, jintasuttisak2022deep}. This shortfall highlights the need for robust, computationally efficient algorithms capable of supporting real-time, on-board UAV processing, which is crucial for numerous field applications.

Finally, large-scale spatial analysis of palms in tropical forests is relatively rare, largely due to limitations in localization methods and the lack of extensive, high-quality datasets. To ensure effective monitoring and conservation, it is crucial to develop robust statistical models that accurately represent the spatial distribution of palms across vast tropical forest areas~\cite{wagner2020regional}. Understanding the ecological mechanisms driving plant distribution is fundamental, especially given the urgent need to address the rapid degradation of wilderness areas in recent decades~\cite{sutherland2013identification, di2019wilderness}.

Addressing the challenges of detecting, segmenting, quantifying, and analyzing the spatial distribution of palms in tropical forests using UAV imagery, this work presents the following main contributions:
\begin{enumerate}
    \item We develop a dataset through extensive fieldwork across 21 sites in western Ecuador, spanning a rainfall gradient from the Choco's wettest forests to the edge of the tropical dry forest at the limit of the Sechura desert (5800mm to 1400mm precipitation). This gradient is reflected in the canopy palm species composition and the canopy characteristics of the forests, with dominant species in the northern, wetter sites being absent in the southern, drier sites, and vice versa. We annotate 1,500 image patches from two sites with bounding boxes, capturing 7,356 palm instances, and additionally mark the landscape center points of 7,603 palms across five sites for counting.
    \item We introduce a flexible framework for evaluating models in real-time palm detection, segmentation, and counting, effective even with limited computational resources. We further enhance the model's interpretability using saliency maps to spotlight critical decision-making areas. Additionally, we investigate the balance between label quantity and model performance, revealing that training set size significantly impacts detection but minimally affects counting performance.
    \item We present a Poisson-Gaussian reproduction algorithm for simulating the spatial distribution of palms. This model combines a Poisson process with a local Gaussian distribution to accurately replicate observed point patterns. The model provides key insights into ecological dynamics and palm population structures across diverse forest environments.

\end{enumerate}

The paper is organized as follows: Section \ref{sec:related} reviews research on object detection, segment anything models (SAMs), palm identification, and spatial point pattern analysis. Section \ref{sec:material} describes our dataset and methodology, including the study area, data collection, preprocessing, and annotation procedures, and the PalmDSNet framework for palm detection, segmentation, and counting, along with the Poisson-Gaussian model for simulating palm distributions. Section \ref{sec:result} presents our experimental design, numerical results, and analysis of palm localization and spatial distribution. Section \ref{sec:conclude} summarizes our findings and proposes future research directions.

\section{Related Work}
\label{sec:related}

\subsection{Object Detection}
\label{sec:object}

Object detection is a computer vision task aimed at identifying and locating objects within digital images or video frames. This task involves not only classifying the objects but also precisely predicting the location of each object through bounding boxes~\cite{zou2023object}. Object detection serves as a fundamental step for more complex tasks such as image segmentation, object tracking, and automated scene understanding, underpinning many applications. Two main approaches that have advanced the field are Detection Transformer (DETR)~\cite{carion2020end, zhao2023detrs} and You Only Look Once (YOLO)~\cite{yolov8, yolov9, yolov10}.

\subsubsection{DETR} 

Recent developments in object detection have been significantly influenced by the introduction of attention-based transformers, such as DETR~\cite{carion2020end}. DETR approaches object detection as a set prediction problem, which eliminates the reliance on non-maximum suppression (NMS) and simplifies the detection pipeline. It employs a transformer-based encoder-decoder architecture, combined with a set-based global loss and bipartite matching, to deliver accurate object predictions. The model utilizes a fixed set of learned object queries to capture and reason about the relationships between objects and their global context within the image.

Real-Time Detection Transformer (RT-DETR)~\cite{zhao2023detrs} enhances the original DETR model by incorporating a convolutional backbone and an efficient hybrid encoder. This adaptation optimizes the processing of multi-scale features through a combination of intra-scale interactions and cross-scale fusion. RT-DETR achieves real-time performance while preserving high accuracy. It also provides flexibility in adjusting inference speed through modifications to decoder layers, without requiring retraining~\cite{zhao2023detrs}.

\subsubsection{YOLO}

YOLO is a real-time object detection framework that reformulates detection as a single regression task, predicting bounding boxes and class probabilities directly from images. This approach enables rapid processing, making YOLO suitable for applications in various real-world scenarios, such as autonomous driving, robotic navigation, and pedestrian tracking~\cite{diwan2023object}. The YOLO series, particularly versions v8 through v10~\cite{yolov8, yolov9, yolov10}, is renowned for its exceptional balance between performance and efficiency.

YOLOv8~\cite{yolov8} introduces a refined architecture that includes a cross-stage partial network for efficient feature extraction, an enhanced path aggregation network for multi-scale feature fusion, and an optimized detection head responsible for predicting object locations and classes in different scales~\cite{wang2020cspnet}. YOLOv9~\cite{yolov9} advances this by incorporating gradient enhanced learning and augmentation network and programmable gradient information to improve training efficiency. YOLOv10~\cite{yolov10} further evolves by eliminating the need for NMS through consistent dual assignments, and features a lightweight classification head, spatial-channel decoupled down-sampling, and a compact inverted block design, which significantly reduces inference latency and makes it the fastest YOLO model to date.

\subsection{Segment Anything Model}
\label{sec:sam}

Segment Anything Models (SAMs) represent a class of advanced segmentation algorithms that can delineate every object within an image, irrespective of its type. Unlike conventional segmentation approaches focused on specific classes, SAMs generalize across varied object types, thus broadening their applicability~\cite{kirillov2023segment, ravi2024sam, mobile_sam, zhang2023comprehensive}. SAMs can segment entire images autonomously or use prompts like points, boxes, or text to direct segmentation. The foundational Segment Anything Model (SAM)~\cite{kirillov2023segment} and its second generation SAM 2~\cite{ravi2024sam}, along with the mobile counterpart Mobile SAM~\cite{mobile_sam}, all demonstrate these capabilities.

\subsubsection{SAM} 

The foundational Segment Anything Model (SAM)~\cite{kirillov2023segment} leverages the SA-1B dataset, which includes over 1 billion masks from 11 million images, to facilitate prompt-based and zero-shot segmentation. SAM integrates a Vision Transformer (ViT)~\cite{dosovitskiy2020image} for image encoding, a prompt encoder for processing input prompts, and a dynamic mask decoder for accurate segmentation across various tasks.

The second generation, SAM 2~\cite{ravi2024sam}, extends these capabilities to video segmentation, allowing for object tracking across frames. Trained on both the SA-1B dataset and the SA-V dataset, which includes 50.9 thousand videos and 642.6 thousand masklets, SAM 2 enhances performance by incorporating a masked autoencoder (MAE) pre-trained Hiera encoder~\cite{he2022masked, ryali2023hiera}. This upgrade enables the use of multiscale features, significantly improving the model's overall effectiveness.

\subsubsection{Mobile SAM} 

Mobile SAM~\cite{mobile_sam} optimizes segmentation for real-time mobile use by employing a simplified image encoder to reduce computational demands. It utilizes a dual-model framework, with a teacher and student model both based on ViT backbones. The encoder's feature map is processed by a prompt-guided mask decoder to generate segmentation masks. Knowledge distillation~\cite{hinton2015distilling} trains the student model, minimizing distillation loss:
$L_d=\alpha \cdot T^2 \cdot \operatorname{MSE}\left(\mathbf{p_{t}}, \mathbf{p_{s}}\right)+(1-\alpha) \cdot \operatorname{MSE}\left(\mathbf{y}, \mathbf{p_{s}}\right),$
where $\operatorname{MSE}$ denotes the mean squared error, $\mathbf{p_{t}}$ are the soft targets from the teacher model at temperature $T$, $\mathbf{p_{s}}$ are the outputs from the student model, $\mathbf{y}$ is the ground truth, $\alpha$ balances the losses between soft targets and ground truth, and $T$ adjusts the influence of the soft targets. This approach compresses the SAM model into a more efficient Mobile SAM, enabling its deployment in mobile and resource-constrained environments.

\subsection{Palm Identification}

Recent advancements in palm identification employ segmentation, classification, and object detection techniques. Segmentation involves dividing an image into polygonal segments that represent different objects or classes; for palm identification, it focuses on isolating palm crowns and isolated palm leaves from the background. These methods typically identify the centers of palm features and create a disc-shaped mask around them as ground truth. For instance,~\citeauthor{gibril2021deep} utilized a U-Net model with residual networks to segment date palms in UAV images of plantations, characterized by regular palm distributions against bare soil backgrounds~\cite{gibril2021deep}. Similarly,~\citeauthor{freudenberg2019large} applied two U-Nets of varying complexities to high-resolution satellite imagery of plantations, where palms are regularly spaced and the background is primarily soil~\cite{freudenberg2019large}. Although the precision and recall in these studies range from 88\% to 94\%, the scenarios are relatively simple due to the clear backgrounds and regular palm distribution.

Classification techniques assign category labels to entire images, individual pixels, or specific objects within images, such as identifying palm segments based on visual characteristics. A common approach in object detection via classification is the sliding window technique, where a fixed-size window moves across the image to classify each sub-region, though this approach is computationally intensive. 
\citeauthor{li2016deep} employed CNNs with sliding windows to detect and count oil palms in QuickBird satellite images, benefiting from regular palm distribution and relatively homogeneous, contrasting backgrounds~\cite{li2016deep}. In~\citeauthor{cui2024palmprobnet}, a probabilistic approach was developed to detect palms in high-resolution UAV images of Ecuadorian rainforest, using dual-scale labeling and varying sliding window sizes to create probability maps for palm presence in dense forests~\cite{cui2024palmprobnet}. These methods consistently achieve over 95\% accuracy in patch classification but require complex postprocessing steps to derive individual palm locations.

Object detection offers a direct approach by using bounding boxes for localization, which demands high-quality annotations, particularly for applications requiring precise and rapid detection. Among the various frameworks available, YOLO is particularly favored for its balance of speed and accuracy~\cite{diwan2023object}. \citeauthor{qin2021ag} introduced Ag-YOLO, a YOLOv3-Tiny variant with focal loss to improve the detection of smaller areca palms in UAV images~\cite{qin2021ag}. Similarly, \citeauthor{jintasuttisak2022deep} used YOLOv5 to detect date palms in the United Arab Emirates, where palms are clearly delineated against the background~\cite{jintasuttisak2022deep}. These methods generally achieve precision and recall rates up to 92\%, though often tested on smaller datasets with relatively simple backgrounds.

\subsection{Spatial Point Pattern Analysis}

\begin{table}[tb]
\centering
\caption{Ripley's \(G\), \(F\), and \(J\) Functions. Here, $\hat{d}_i$ represents the nearest neighbor distance for each observed point $i$, and $\Tilde{d}_j$ is the nearest neighbor distance from a simulated point $j$ to the observed points. $\mathbbm{1}(\cdot)$ is an indicator function.}
\label{tab:ripley_gfj}
    \begin{tabular}{cc}
        \toprule
        \textbf{Function} & \textbf{Formulation} \\
        \midrule
        $G$ & $G(d) = \frac{1}{N_o}\sum_{i=1}^{N_o} \mathbbm{1}(\hat{d}_i < d)$, $N_o = $ observed events\\
        \\[0.1pt]
        $F$ & $F(d) = \frac{1}{N_r}\sum_{j=1}^{N_r} \mathbbm{1}(\Tilde{d}_j < d)$, $N_r = $ simulated events \\
        \\[0.1pt]
        $J$ & $J(d) = {(1-G(d))}/{(1-F(d))}$ \\
        \bottomrule
    \end{tabular}
\end{table}

In forest ecosystems, dynamic ecological processes such as seed dispersal, species competition, and mortality often shape certain spatial patterns for trees that can be detected and modeled~\cite{Petritan2014forest}. A robust analysis of these spatial distributions is crucial for understanding forest structure and its ecological and evolutionary dynamics~\cite{Gadow2012structure, May2015Moving, Jia2016MechanismUT}. One natural approach to spatial distribution analysis involves representing each tree in the study area as a point, and then investigating both (1) the characteristics of the spatial pattern -- what the pattern looks like, and (2) the underlying point process that generates the observed data -- how the pattern arises.

To address the first question -- characterizing point patterns -- Ripley's set of statistical functions, often referred to as Ripley’s alphabet, is widely employed~\cite{Ripley1981, Baddeley2016}. Among these, three key functions, $G$, $F$, and $J$, stand out for their simplicity and effectiveness. The $G$ function, $G(d)$, calculates the proportion of nearest neighbor distances within the sample that are less than or equal to a given distance $d$, thereby providing insights into the degree of clustering or dispersion within the data. The $F$ function, in contrast, mirrors the $G$ function but measures the nearest neighbor distances from a set of randomly distributed points to those in the observed pattern, effectively serving as a measure of dispersion between the observed pattern and a completely random one. The $J$ function combines the information from both the $G$ and $F$ functions, offering a more comprehensive analysis by simultaneously considering intra-pattern and inter-pattern spatial relationships. By applying these Ripley’s functions, one can derive clear and concise summary statistics that effectively describe the attributes of point pattern data. Table \ref{tab:ripley_gfj} provides the mathematical definitions of each of these functions.

In addition to functions that consider only the nearest neighbor for each point, there are widely used alternatives designed to detect spatial patterns at multiple scales, such as the $K$ function~\cite{Ripley1981, K_L_func1, Haase1995} and its derivative, the $L$ function~\cite{Besag1977}. These functions, due to their mathematical properties, are highly effective for analyzing multivariate and multi-scale spatial patterns. However, their application is computationally intensive and requires comprehensive data collection across large areas~\cite{Goreaud2000, ben2021spatial}. Since our research is focused on a specific and confined region of palm distribution, we prioritize single-scale functions to optimize efficiency while still capturing relevant spatial patterns.

In simulating the process that generates the observed spatial patterns, it is advantageous to employ mathematical point process models that incorporate stochastic mechanisms. The homogeneous Poisson process, widely used in ecological studies~\cite{Velázquez2016evaluation, Wiegand2004RingsCA}, assumes that points are randomly and independently distributed within the observation window, leading to point patterns that exhibit complete spatial randomness with no spatial trends or associations among points. However, this simplistic model often fails to capture the complexity of real-world environmental conditions. To address this limitation, more advanced models have been developed. In contrast to the homogeneous Poisson process which assumes a constant intensity, the heterogeneous Poisson process employs a spatially varying intensity function, enabling a closer approximation of real-world data distributions~\cite{Wiegand2004RingsCA, Renner2015HP, Carrer2018HP}. Additionally, the Thomas cluster process, a widely adopted model, introduces Poisson cluster processes across multiple scales, thereby enhancing the model's capacity to depict the intricate structures observed in empirical data~\cite{Law2009thomas, Jácome2016thomas}.


\begin{figure*}[t]
    \centering
    \includegraphics[width=\linewidth]{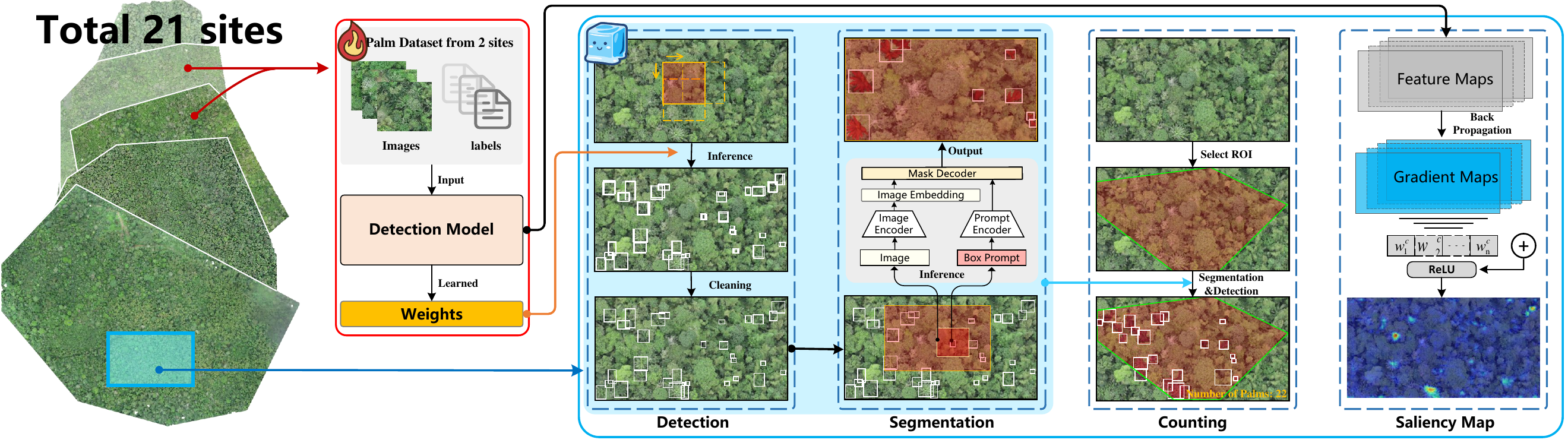}
    \caption{Workflow of PalmDSNet. The model is trained on labeled patches for accurate palm detection, with the refined weights then used to infer bounding boxes on new images. These boxes guide the segmentation and counting processes across selected regions or landscape scales. Saliency maps are also produced to enhance the interpretability of palm localization.}
    \label{fig:workflow}
\end{figure*}

\section{Materials and Methods}
\label{sec:material}

\subsection{Dataset}
\label{sec:data}

\subsubsection{Study Sites and Data Collection}

The data for this study come from tropical forest in western Ecuador Chocó centered on the Fundación para la Conservación de los Andes Tropicales Reserve and adjacent Reserva Ecológica Mache-Chindul park (FCAT; 00$^\circ$23'28'' N, 79$^\circ$41'05'' W), and the Jama-Coaque Ecological Reserve (00$^\circ$06'57'' S, 80$^\circ$07'29'' W). The FCAT reserve is high diversity humid tropical forest at $\sim$500 m elevation, receiving $\sim$3000 mm precipitation yr$^{-1}$ accompanied by persistent fog during the period of lower precipitation. The Jama-Coaque reserve spans from the boundary of tropical moist deciduous/tropical moist evergreen forest at the lower elevations ($\sim$1000 mm precipitation yr$^{-1}$, $\sim$250 m asl) to fog-inundated wet evergreen forests above 580m to 800m. These forests harbor several regularly occurring palm species with canopy-exposed crowns~\cite{browne2016diversity, lueder2022functional}, with the dominant species being \textit{Iriartea deltoidea} and \textit{Socratea exorrhiza} at FCAT and \textit{Astrocaryum standleyanum} and \textit{Socratea exorrhiza} at Jama-Coaque.

To capture detailed spatial data, UAV operations were conducted in two phases across 96 plots in 21 areas, covering a total of 1995 hectares with a ground spacing distance (GSD) $<$6 cm. The first phase, in June 2022, involved flying over 95 hectares using a DJI Phantom 4 RTK drone equipped with a 1'' CMOS sensor. Mission planning was done using GS RTK, with flights conducted at an altitude of 90 meters above ground level (AGL) with 70\% sidelap and 80\% frontlap, resulting in 387 photos processed with Agisoft Metashape 2.0 to produce orthomosaics. An orthomosaic is a large, georeferenced image created through photogrammetry, compiled from multiple overlapping images of a specific geographic area. These images are orthorectified to standardize perspective and spatial scale while embedding spatial metadata. However, localized distortions may occur, particularly along the edges due to the orthorectification process. The second phase, conducted in February 2023, expanded coverage to 1900 hectares, flying at 150 meters AGL and capturing 8458 photos using the same drone and software. Subsequent data processing involved noise reduction, edge trimming of orthomosaics, and the generation of Digital Surface Models and Digital Terrain Models.

\subsubsection{Manual Labels}

To develop a high-quality training and validation dataset for our models, we manually labeled drone imagery from two sites within the FCAT Reserve. We extracted 800$\times$800 patches from orthomosaics, selecting a subset of 1,500 patches representing various image qualities and palm densities. Precise bounding boxes were annotated around palm crowns and isolated leaves. Additionally, for counting purposes, we manually labeled five orthomosaics using ArcGIS Pro 3.3.1, systematically placing georeferenced points at the centers of palm crowns visible in the canopy imagery.

Unlike studies focused on regularly spaced plantation palms~\cite{gibril2021deep, li2016deep, jintasuttisak2022deep}, our research examines natural forest data (see Figure \ref{fig:manual}). The manual annotation of bounding boxes was particularly hard due to palms overlapping with one another and other trees, and the occurrence of artifacts and distortion from orthomosaic processing. Similarly, labeling the center points of palms across five sites was labor-intensive but essential for conducting landscape-level analysis and validating our model's counting accuracy against expert assessments.

\subsection{Palm Detection and Segmentation Network}
\label{sec:method}

This section introduces the Palm Detection and Segmentation Network (PalmDSNet), a framework specifically designed to detect, segment, and count palms in the dense and diverse tropical forests (see Figure \ref{fig:workflow}). Unlike previous approaches that either convert detection into segmentation~\cite{gibril2021deep, freudenberg2019large}, treat it as a classification task~\cite{li2016deep, cui2024palmprobnet}, or focus solely on detection without segmentation~\cite{qin2021ag, jintasuttisak2022deep}, our approach employs a state-of-the-art detection model enhanced with an integrated segmentation head. Additionally, our framework tackles the complexities of dense rainforests with irregular palm distribution and overlapping vegetation, demonstrating the ability to locate palms even from a single leaf, by applying it to a dataset containing these challenging images.

PalmDSNet is a flexible architecture that allows for substitution of different detection and segmentation methods to suit various contexts. We employ fine-tuned backbones for palm detection, followed by segmentation models to delineate the detected palms, visualizing their spatial distribution. The detection results are refined for precise counting, which is displayed alongside segmentation maps to provide a comprehensive overview. Moreover, we enhance result interpretability using saliency maps, which highlight the focal areas influencing the detection network's decisions. This multi-phase approach combines state-of-the-art object detection and segmentation models to achieve accurate, real-time palm counting.

\subsubsection{Palm Detection} 

We train networks to locate palms within forest landscapes depicted in orthomosaic images, which are segmented into patches for detailed analysis. These networks are designed to be flexible, allowing for the substitution of alternative models, such as YOLOs or DETRs~\cite{carion2020end, zhao2023detrs, yolov8, yolov9, yolov10}, based on specific analytical needs. Training involves various image augmentations, such as adjustments in hue, saturation, and brightness, as well as rotations, scaling, translations, and flips, to enhance model robustness across diverse environments. During inference, each patch is analyzed to locate palms, with their coordinates recorded for further processing. NMS is employed to eliminate redundant detections from overlapping patches, ensuring that each identified palm is uniquely processed in the subsequent segmentation phase.

\subsubsection{Palm Segmentation} 

Segmentation is performed during inference due to the absence of segmentation masks in the training dataset. Detected palms serve as centers, with their surrounding areas included for contextual information. The bounding box from the detection phase acts as a prompt for a chosen SAM, guiding it to accurately segment the palm. This method is particularly efficient for focusing on specific regions within extensive forest landscapes.

\subsubsection{Palm Counting} 

Palm counting starts by defining the scope of analysis, which can be the entire landscape or a specific region of interest (ROI). For ROIs, a user-drawn polygon is enclosed within a rectangular bounding box to ensure full coverage during detection and segmentation. The detection module identifies individual palms and logs their coordinates, followed by redundancy removal using NMS. The segmentation branch then generates masks for these palms based on detection outputs. These masks, along with the bounding boxes, are visualized to enable thorough analysis and verification of detected palms within the ROI.

\subsubsection{Saliency Map}

To enhance model interpretability and reliability, we use Grad-CAM~\cite{selvaraju2017grad, jacobgilpytorchcam}. This technique creates a coarse localization map by leveraging gradients flowing into the final convolutional layer, highlighting the regions of the image most crucial for predicting the target concept. Grad-CAM provides visual explanations for the model's predictions, revealing key areas of focus and ensuring that the network emphasizes relevant features for accurate palm localization.

\subsection{Poisson-Gaussian Palm Reproduction Algorithm}

\begin{algorithm}[t]
\SetAlgoLined
\Input{$\mathbf{p}$ (list of candidate $p$), $\boldsymbol{\sigma}$ (list of candidate $\sigma$), $X$ (set of observed palm coordinates), $N$ (number of Simulations)}
\Output{$p^*, \sigma^*$ (optimal $p$ and $\sigma$ that minimizes $d$)}

\BlankLine
Initialize $p^* = 0$, $\sigma^* = 0$, $d_{min} = \infty$, $\hat{X} = \emptyset$\;
Compute Ripley's functions $\mathbf{g} = G(X)$ and $\mathbf{f} = F(X)$\;

\BlankLine
\SetKwFunction{FSimulate}{Simulate}
\SetKwProg{Fn}{Procedure}{:}{}
\Fn{\FSimulate{$p, \sigma$}}{
    Initialize $\hat{X}$ with a random 2D point generated uniformly across the spatial extent\;
    \While{$|\hat{X}| < |X|$}{
        Select a random parent palm $\mathbf{x}$ from $\hat{X}$ and generate a random number $p_r$ from $[0, 1]$\;
        \eIf{$p_r < p$}{
            Generate offspring palm using a Gaussian distribution $\mathcal{N}(\mathbf{x}, \boldsymbol{\Sigma})$ around the parent $\mathbf{x}$\, with $\boldsymbol{\Sigma} = [\sigma^2,0;0,\sigma^2]$;
        }{
            Generate offspring palm from a 2D uniform distribution\;
        }
        Append offspring palm to $\hat{X}$\;
    }
    \KwRet $\hat{X}$\;
}

\BlankLine
\ForEach{$p$ in $\mathbf{p}$}{
    \ForEach{$\sigma$ in $\boldsymbol{\sigma}$}{
        Initialize $d = 0$\;
        \For{$i = 1$ \KwTo $N$}{
            $\hat{X}_i = \FSimulate(p, \sigma)$\;
            Compute $\mathbf{g_s}_i = G(\hat{X}_i)$ and $\mathbf{f_s}_i = F(\hat{X}_i)$\;
            Integrate the difference for the $i$-th trial: 
            $d_i = \int_{x} \left| \mathbf{g} - \mathbf{g_s}_i \right| \, dx + \int_{x} \left| \mathbf{f} - \mathbf{f_s}_i \right| \, dx$\;
            $d = d + d_i$\;
        }

        \If{$d < d_{min}$}{
            Update $d_{min} = d, p^* = p$, $\sigma^* = \sigma$\;
        }
    }
}
\KwRet $p^*, \sigma^*$\;

\caption{Poisson-Gaussian Palm Reproduction}
\label{alg:optimize_palm}
\end{algorithm}

\label{sec:optimize_palm}

This section introduces the proposed Poisson-Gaussian model to simulate palm spatial reproduction that optimizes the reproduction parameters to minimize the discrepancy between observed and simulated point patterns of palm coordinates (see Algorithm \ref{alg:optimize_palm}). By utilizing this model, we aim to gain insights into both global and local aspects of palm reproduction, including long-range and short-range reproductive dynamics. Understanding the spatial distribution of palms within tropical forests can further reveal the relationships among different palm species and individuals of the same species, ultimately helping to determine whether palms compete for resources or can coexist in close proximity.

The proposed Poisson-Gaussian model integrates a Poisson point process~\cite{kingman1992poisson} with a local Gaussian distribution to simulate the dual nature of palm propagation. The Poisson point process captures the random spread of palms, such as through animal-mediated dispersal or other stochastic processes, leading to a more uniform distribution of individuals. The Gaussian distribution, on the other hand, represents the tendency of seeds to fall and germinate near their parent trees, reflecting a more localized clustering effect. By adjusting these processes to align with observed spatial patterns, the model provides insights into the ecological dynamics and spatial organization of palm populations in tropical forests. This proposed approach is preferable to the Student's $t$ distribution~\cite{clark1999seed}, as it generates fewer local samples while allowing a portion of samples to disperse globally, thus better aligning with the concept of negative density dependence (NDD)~\cite{metz2010widespread}. NDD posits that as population density increases, the probability of seed germination and survival decreases, encouraging species coexistence and spatial distribution patterns more reflective of ecological realities.

The optimization process begins with initializing \( p^* \) and \( \sigma^* \) to zero and setting \( d_{min} \) to infinity. Ripley's function \( \mathbf{g} \) and the empirical distribution function \( \mathbf{f} \) are computed for the observed set \( X \), serving as benchmarks for comparison. The \texttt{Simulate} function then generates synthetic point patterns based on candidate values of \( p \) and \( \sigma \). The simulation starts with a single randomly placed point and probabilistically adds new points: with probability \( p \), offspring are drawn from a Gaussian distribution centered around a randomly selected parent, and with probability \( 1-p \), offspring are uniformly distributed. This process continues until the synthetic set \(\hat{X}\) matches the size of \( X \).

For each parameter combination, $N$ simulations are performed to compute the total discrepancy. Simulated patterns are compared to the observed data using Ripley's function and the empirical distribution function. The discrepancy \( d_i \) for each trial is measured as the integral of absolute differences between observed and simulated functions. The total discrepancy \( d \) is the aggregate of discrepancies across all trials, with the optimal parameters identified as those minimizing \( d \). This optimization framework effectively balances local clustering with global dispersion patterns to best replicate the observed spatial distribution of palms.

\begin{figure*}[t]
    \centering
    \includegraphics[width=\linewidth]{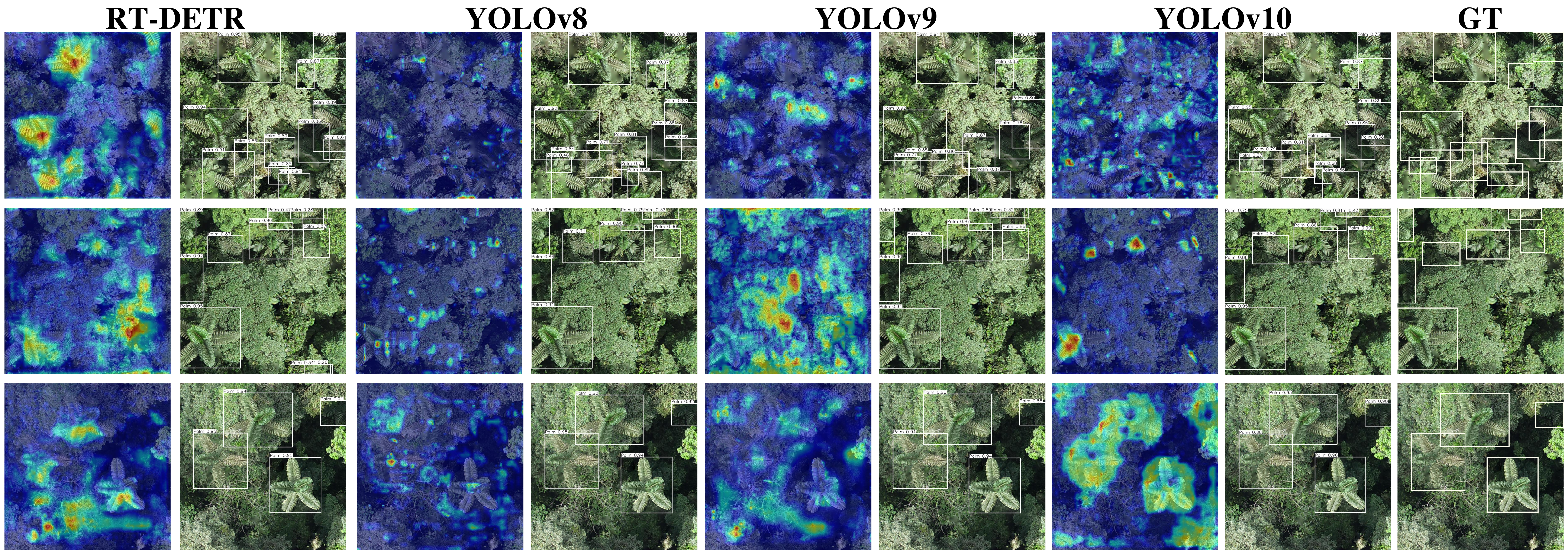}
    \caption{Comparative Visualization of Detection Outcomes and Saliency Maps Across Different Models in PalmDSNet. Each pair of columns displays detection results with bounding boxes and confidence levels alongside saliency maps derived from detection model weights.  The final column showcases the labeled ground truth (GT).}
    \label{fig:comparison}
\end{figure*}

\section{Experimental Results}
\label{sec:result}

This section outlines the performance evaluation of our PalmDSNet framework, systematically addressing several research objectives across different ecological sites. Initially, we investigate the performance of various detection models, including RT-DETR~\cite{zhao2023detrs}, YOLOv8~\cite{yolov8}, YOLOv9~\cite{yolov9}, and YOLOv10~\cite{yolov10} under diverse training conditions to understand their accuracy and efficiency. Saliency maps~\cite{selvaraju2017grad} are ultilized to help understand the focus regions of different models for palm detection in densely populated tree crowns. Following detection, we assess the capabilities of segmentation models SAM~\cite{kirillov2023segment}, SAM 2~\cite{ravi2024sam}, and Mobile SAM~\cite{mobile_sam}, focusing on their inference time and the visual quality of segmentation results. We then explore the impact of training data volume on the accuracy of these models and evaluate their counting performance, a critical component for ecological analysis. Finally, we apply these models to orthomosaics to assess their efficiency in real-time, large-scale landscape analysis.

Additionally, we evaluate our proposed bimodal palm reproduction model, which combines a Poisson point process with a local Gaussian distribution. This model simulates the spatial distribution of palms by accounting for both random spread and localized clustering effects. By comparing simulated patterns with observed data, we assess how well the model replicates real-world palm distribution. This evaluation provides valuable insights into the ecological dynamics of palm populations, including interspecies interactions and habitat coexistence, enhancing our understanding of palm community structuring in tropical forests.

\begin{table*}[ht]
\centering
\caption{Performance Comparison of Detection Models for PalmDSNet.}
\label{tab:detection}
\resizebox{\textwidth}{!}{%
\begin{tabular}{cccccccccccccc}
\toprule
\multirow{2}{*}{\textbf{Model}} & \multirow{2}{*}{\textbf{GFLOPS}} & \multirow{2}{*}{\textbf{Params (M)}} & \multirow{2}{*}{\textbf{Latency (ms)}} & \multicolumn{5}{c}{\textbf{Validation (10\% Training)}} & \multicolumn{5}{c}{\textbf{Validation (90\% Training)}} \\
\cmidrule(lr){5-9} \cmidrule(lr){10-14}
 & & & & \textbf{Precision} & \textbf{Recall} & \textbf{$\text{AP}_{50}$} & \textbf{$\text{AP}_{75}$} & \textbf{mAP} & \textbf{Precision} & \textbf{Recall} & \textbf{$\text{AP}_{50}$} & \textbf{$\text{AP}_{75}$} & \textbf{mAP} \\
\midrule
RT-DETR & 222.5 & 65.5 & 22.4 & 0.845 & 0.778 & 0.843 & 0.560 & 0.532 & \textbf{0.923} & 0.905 & \textbf{0.955} & \textbf{0.822} & \textbf{0.734} \\
YOLOv8  & 257.4 & 68.1 & 22.9 & 0.859 & 0.773 & 0.849 & 0.569 & 0.532 & 0.902 & \textbf{0.919} & 0.944 & 0.725 & 0.672 \\
YOLOv9  & 189.1 & 57.4 & 24.3 & \textbf{0.869} & \textbf{0.795} & \textbf{0.878} & \textbf{0.640} & \textbf{0.590} & 0.897 & 0.914 & 0.950 & 0.769 & 0.680 \\
YOLOv10 & \textbf{169.8} & \textbf{31.6} & \textbf{21.2} & 0.838 & 0.781 & 0.846 & 0.562 & 0.529 & 0.912 & 0.891 & 0.938 & 0.735 & 0.662 \\
\bottomrule
\end{tabular}
}
\end{table*}

\begin{table*}[ht]
\centering
\caption{Effectiveness of Different Configurations for PalmDSNet Across 5 Sites. Sites marked with $^\star$ are training locations.}
\label{tab:effectiveness}
\resizebox{0.95\textwidth}{!}{%
\begin{tabular}{@{}ccccccccccc@{}}
\toprule
\multirow{2}{*}{\textbf{Site}} & \multirow{2}{*}{\textbf{Size (ha)}} & \multirow{2}{*}{\textbf{No. of Counts}} & \multicolumn{4}{c}{\textbf{Counting Accuracy (10\% Training)}} & \multicolumn{4}{c}{\textbf{Counting Accuracy (90\% Training)}} \\ \cmidrule(lr){4-7} \cmidrule(lr){8-11}
 & & & \textbf{RT-DETR} & \textbf{YOLOv8} & \textbf{YOLOv9} & \textbf{YOLOv10} & \textbf{RT-DETR} & \textbf{YOLOv8} & \textbf{YOLOv9} & \textbf{YOLOv10} \\ \midrule
FCAT 1$^\star$& 21.7  & 284   & \textbf{0.997} & 0.979 & \textbf{0.997} & 0.993 & 0.989 & \textbf{0.997} & \textbf{0.997} & 0.993 \\
FCAT 2$^\star$& 119.4 & 2,569 & \textbf{0.998} & 0.988 & 0.992 & 0.994 & \textbf{0.999} & 0.998 & 0.998 & 0.997 \\
FCAT 3{ }{ }  & 103.1 & 2,206 & \textbf{0.998} & 0.995 & 0.993 & 0.995 & \textbf{0.999} & 0.998 & 0.998 & 0.995 \\
Jama-Coaque 1 & 112.2 & 732   & \textbf{0.969} & 0.876 & 0.872 & 0.862 & \textbf{0.975} & 0.870 & 0.909 & 0.833 \\
Jama-Coaque 2 & 92.3  & 1,596 & \textbf{0.938} & 0.776 & 0.812 & 0.775 & \textbf{0.940} & 0.743 & 0.824 & 0.791 \\
\bottomrule
\end{tabular}%
}
\end{table*}

\subsection{Experimental Setting}

To assess the performance of both PalmDSNet and our bimodal reproduction model, we conducted a series of experiments to ensure robustness and reliability. For PalmDSNet, two distinct data allocation strategies were tested: one using 10\% of the data for training over 100 epochs to evaluate performance with limited samples, and another using 90\% of the data for training over 300 epochs to maximize data utilization. Both trained models were then used to validate the detection and counting performance. Images were cropped to 800$\times$800 pixels, providing a broader receptive field, with a stride of 400 for efficient inference on large orthomosaic images. The model was configured to detect a single class to simplify the architecture and reduce computational requirements. While four RTX 3090 GPUs were utilized for training, only one was used during validation and testing to simulate the constraints of real-time operational settings. 

For the Poisson-Gaussian palm reproduction model, we conducted simulations on a CPU, specifically an Intel Xeon Silver 4210 with 40 cores running at 2.20 GHz, to evaluate its ability to replicate the spatial distribution of palms. Each simulation was performed 10 times (\(N=10\)) to ensure that the results were not influenced by random variation. The Ripley's functions, which quantify spatial patterns, were computed at sampled x-axis coordinates. The trapezoidal rule~\cite{hildebrand1987introduction} was then employed to interpolate these functions and integrate the absolute differences between observed and simulated distributions.

Our evaluation utilized a series of standard metrics for a thorough assessment of the model's performance. Precision and Recall measure the model's accuracy and error rates in detecting objects. Average Precision (AP) at various IoU thresholds quantifies precision across different recall levels, and Mean Average Precision (mAP) aggregates these measures for a comprehensive performance overview. We also examine computational demands by comparing Giga Floating Point Operations Per Second (GFLOPS), the number of parameters (Params), and latency to evaluate efficiency during training and real-time inference. We validate the model's counting ability by comparing its performance on five sites against that of human experts. For the bimodal reproduction model, we used the minimum integrated difference between observed and simulated point patterns to determine the optimal parameter pair $p^*$ and $\sigma^*$.

\subsection{Performance Evaluation of PalmDSNet}

\subsubsection{Detection}

The comparative analysis of detection models for PalmDSNet are encapsulated in Table \ref{tab:detection}, highlighting model size, inference latency, and performance metrics under two distinct training scenarios: 10\% and 90\% of the data used for training. The evaluated models demonstrate diverse computational costs and inference efficiencies. YOLOv10 emerges as the lightest with the fewest parameters and lowest GFLOPS, whereas YOLOv8 is the heaviest. Despite the differences in model size, all models maintain consistent inference times, processing at speeds exceeding 40 frames per second, thus ensuring their capability for real-time applications and suitability for on-board deployment.

Performance analysis reveals that more training data consistently enhances model performance in palm localization. With 90\% of the data allocated for training, there is a noticeable improvement in Precision, Recall, and the quality of bounding boxes, as reflected in higher APs across thresholds. Notably, with only 10\% of the data for training, YOLOv9 leads in detection performance, achieving an mAP of 0.59 and an $\text{AP}_{75}$ of 0.64 -- both over 6\% higher than the nearest competitor. Its precision and recall also exceed the next best by more than 1\%. Conversely, when trained with 90\% of the data, RT-DETR excels, notably in mAP and $\text{AP}_{75}$, reaching 0.734 and 0.822, respectively, each about 5\% higher than its closest competitor. These results suggest that ViT-based methods significant benefits from larger training sets, particularly in bounding box quality.

The visual comparison in Figure~\ref{fig:comparison} using models trained with 90\% data demonstrates their performance across varied scenarios. In the first row of Figure~\ref{fig:comparison}, YOLOv9 struggles with densely overlapped palm crowns, indicating difficulties in complex canopy textures, which is reflected in the saliency map focusing on non-palm tree crowns and leading to missed detections. In the second row, RT-DETR, while showing a slightly higher confidence in palm detection, produces two false positives in the bottom right corner due to a low confidence threshold set to improve recall. Its saliency map confirms a heightened focus in this area. Conversely, YOLOv8 exhibits a similar issue with a false alert on the top right, and YOLOv10, despite its precision, fails to accurately bound a palm leaf there. In the third row, all models perform robustly in a more straightforward setting, correctly identifying palms and proving their efficacy. The saliency maps reveal prioritization of palm leaves and centers, with centers often highlighted more intensely; however, they also display activities in non-palm zones, illustrating that while saliency maps guide model focus, the final detection head -- comprising additional layers responsible for final decision-making -- ultimately confirms the presence of palms, which shows the disparity between model attention and actual detection.

\begin{figure*}[tb]
  \centering
  \begin{subfigure}{0.24\textwidth}
    \includegraphics[width=\textwidth]{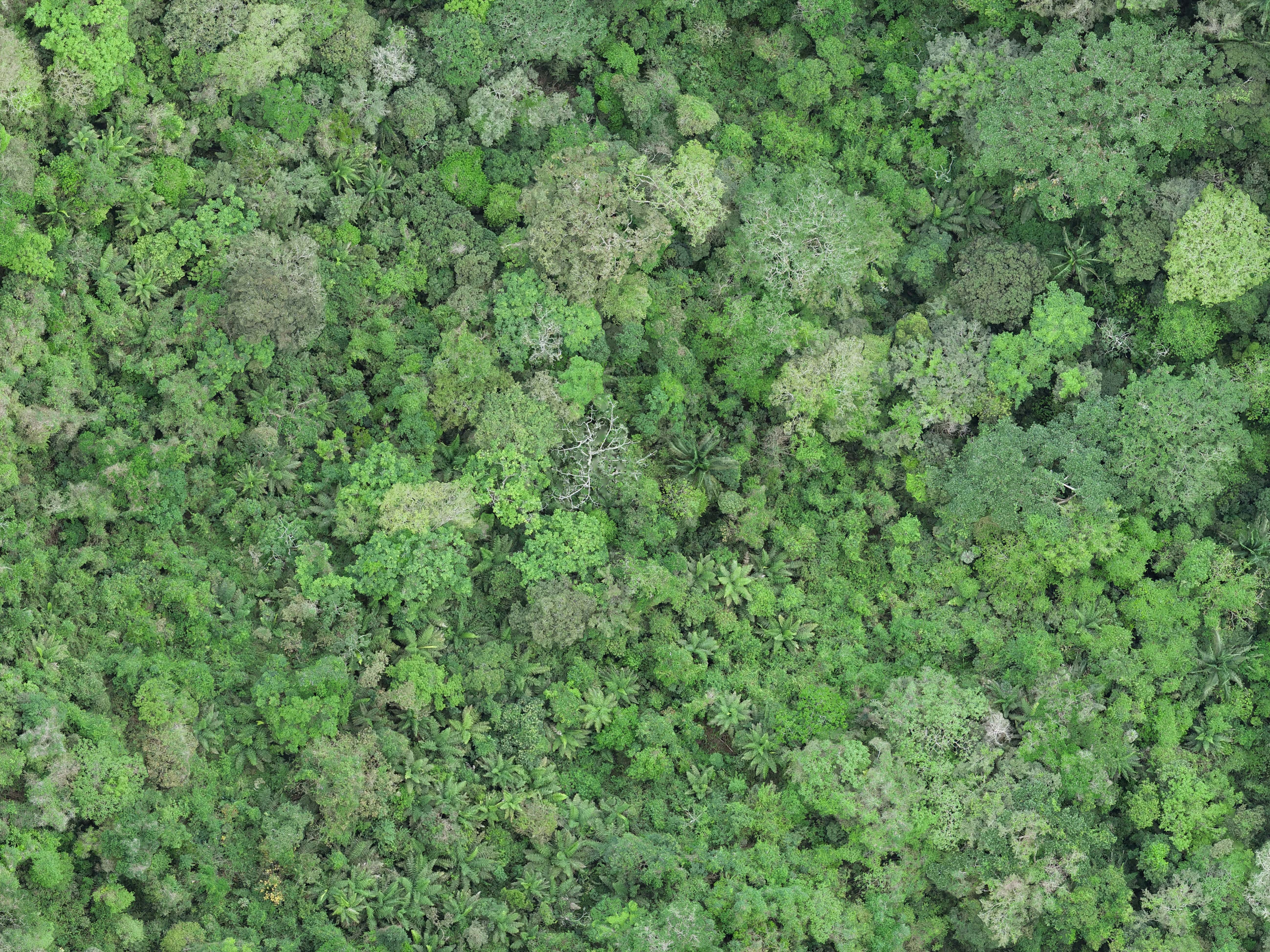}
    \caption{Sample Region}
  \end{subfigure}%
  \hfill
  \begin{subfigure}{0.24\textwidth}
    \includegraphics[width=\textwidth]{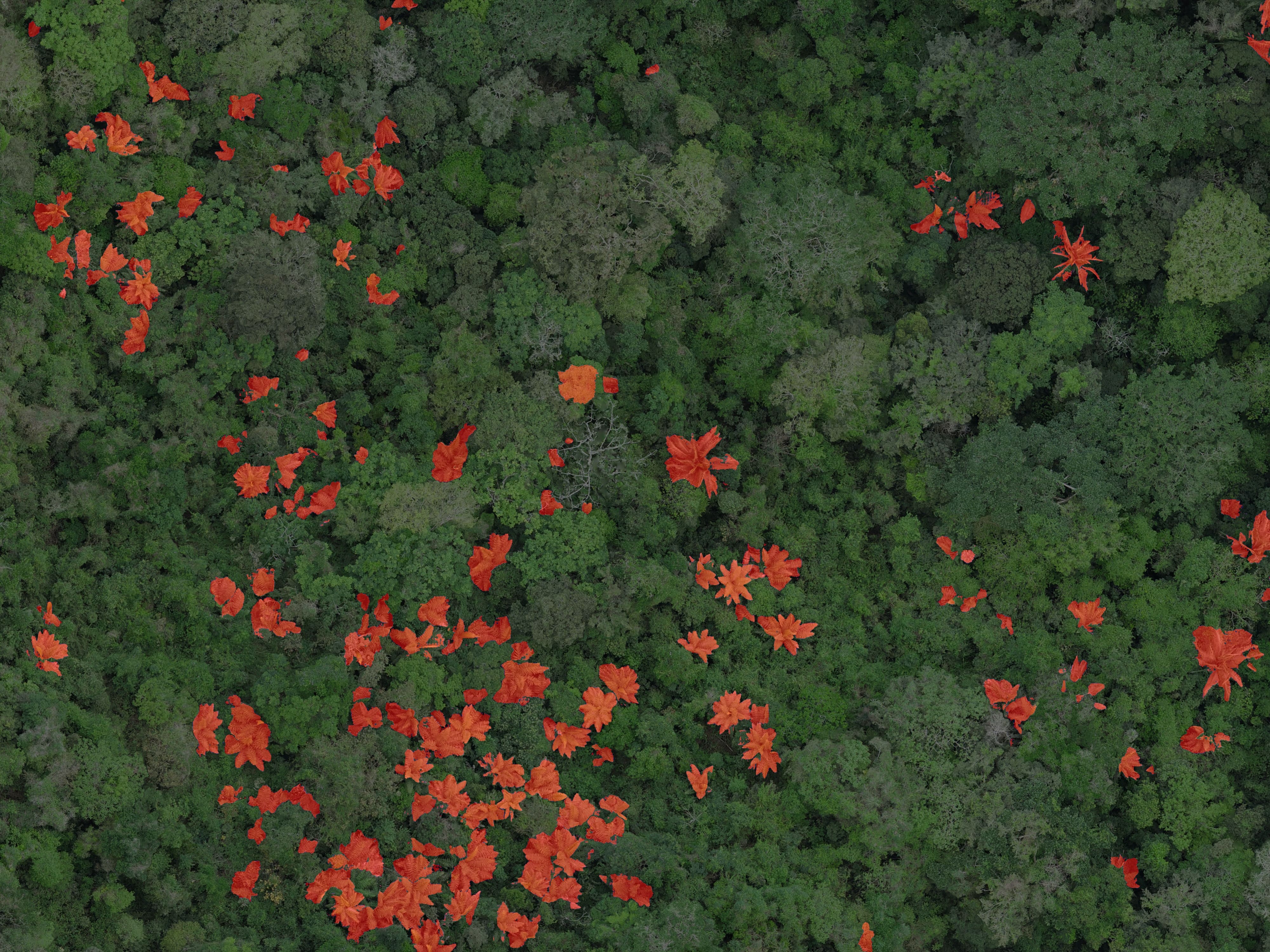}
    \caption{SAM}
  \end{subfigure}%
  \hfill
  \begin{subfigure}{0.24\textwidth}
    \includegraphics[width=\textwidth]{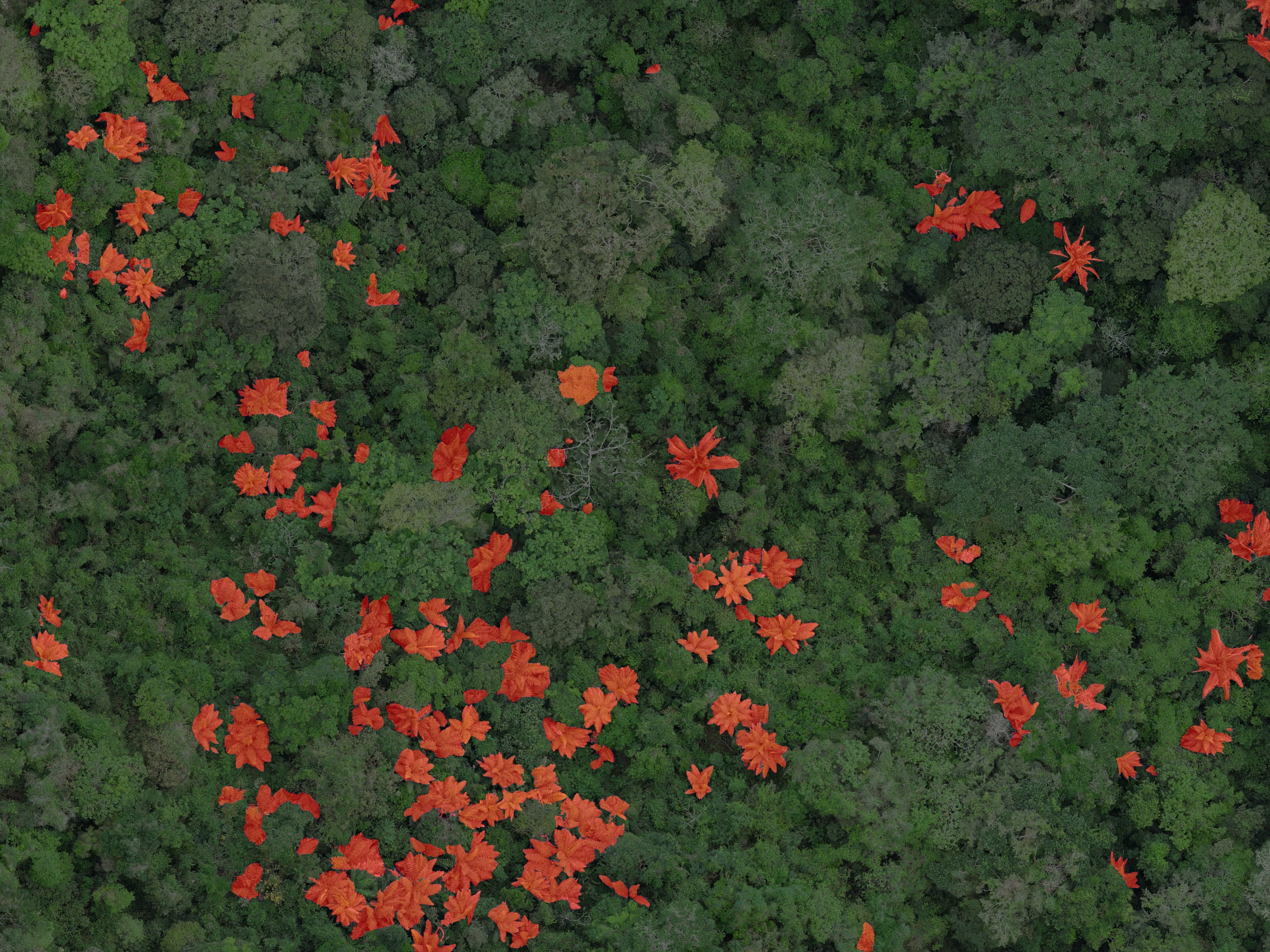}
    \caption{SAM 2}
  \end{subfigure}%
  \hfill
  \begin{subfigure}{0.24\textwidth}
    \includegraphics[width=\textwidth]{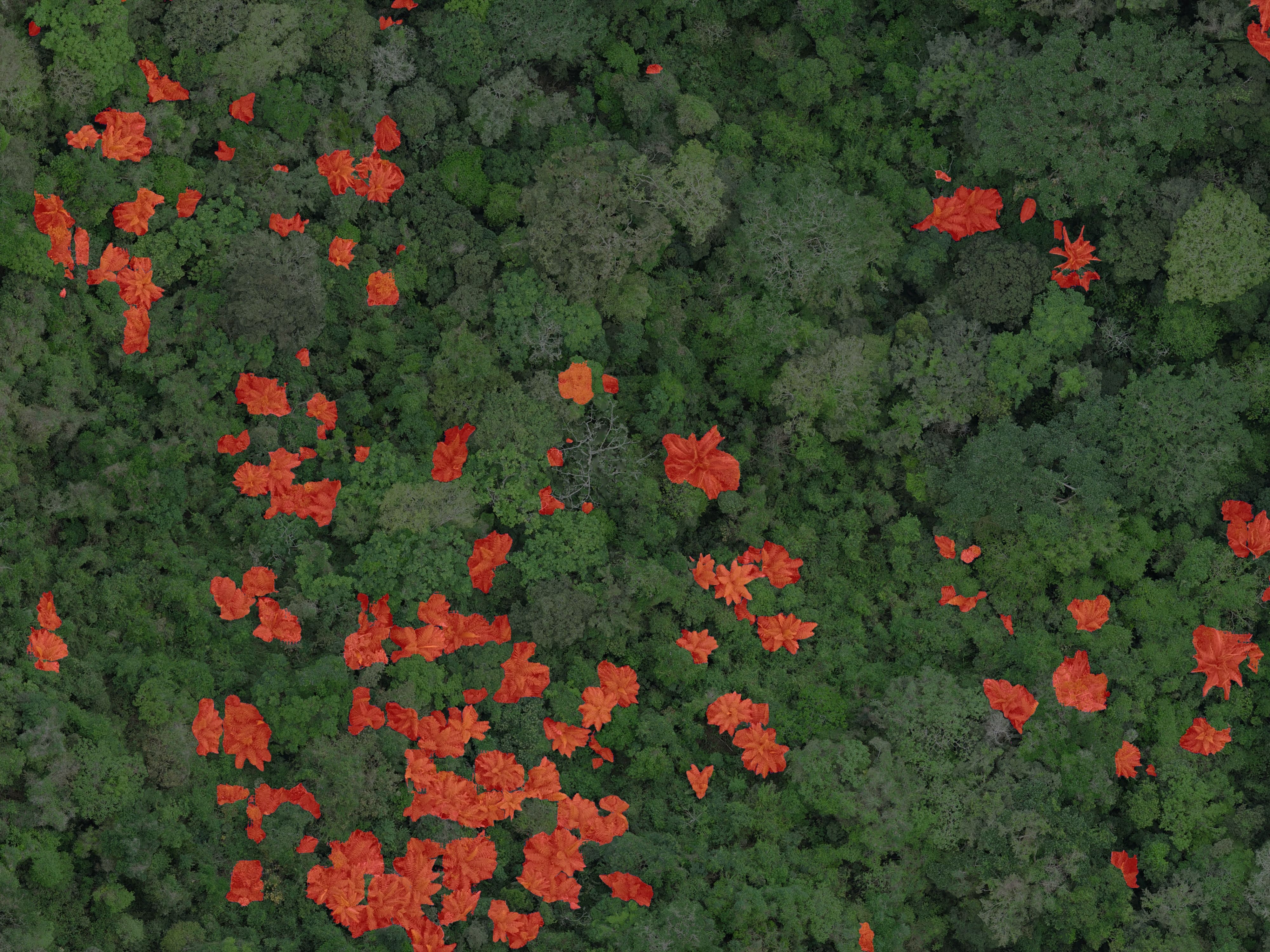}
    \caption{Mobile SAM}
  \end{subfigure}
  \caption{Segmentations of SAM and Mobile SAM.}
  \label{fig:sam}
\end{figure*}

\subsubsection{Segmentation}

The segmentation outcomes from SAM, SAM 2, and Mobile SAM, compared in Figure \ref{fig:sam}, reflect their robust out-of-sample predictive performance in Jama-Coaque 1, a site in another reserve that is significantly different in rainfall, palm species, and forest structure from the environment of training samples. This demonstrates the adaptability of the detection model, despite some expected false alerts due to the set low confidence level in palm detection. SAM excels in boundary precision that closely matches the actual contours of palm leaves, though it may fragment palm crowns into multiple segments where they overlap with other tree crowns. SAM 2, however, outperforms its predecessor by delivering better boundary segmentation while reducing the occurrence of small, broken segments that appears in SAM. At the same time, SAM 2 maintains a high level of segmentation performance, providing a more cohesive representation of palm crowns. In contrast, Mobile SAM includes broader areas as foreground, resulting in more unified palm segmentation but with less precise boundaries. All models perform commendably without supervision and demonstrate robust inference capabilities.

\begin{figure}[t]
    \centering
    \begin{subfigure}{0.7\linewidth}
    \includegraphics[width=\textwidth]{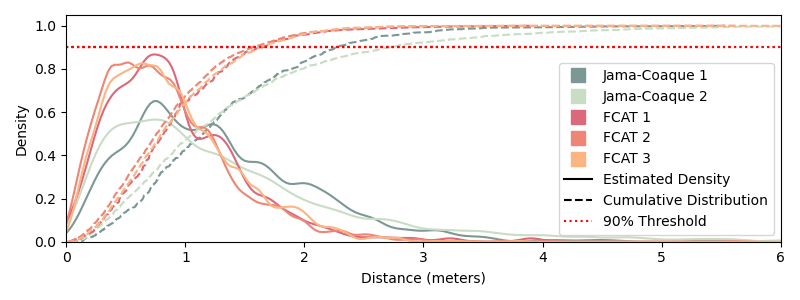}
    \end{subfigure}%
    
    \resizebox{0.6\linewidth}{!}{%
        \begin{tabular}{cccccc}
            \toprule
            \textbf{Statistics} & \textbf{FCAT 1} & \textbf{FCAT 2} & \textbf{FCAT 3} & \textbf{Jama-Coaque 1} & \textbf{Jama-Coaque 2} \\
            \midrule
            Mean (m) & 0.90 & 0.84 & 0.84 & 1.20 & 1.30 \\
            Median (m) & 0.77 & 0.73 & 0.74 & 1.06 & 1.05 \\
            Standard Deviation (m) & 0.60 & 0.54 & 0.51 & 0.78 & 1.02 \\
            \bottomrule
        \end{tabular}
    }
    \caption{Distribution of Distance Shifts Across Sites.}
    \label{fig:kde}
\end{figure}

\begin{table}[t]
\centering
\caption{Latency of Different Configurations for PalmDSNet Across 5 Sites. Segmentation latency is measured based on the detection results of RT-DETR.}
\label{tab:latency}
\resizebox{0.55\linewidth}{!}{%
\begin{tabular}{@{}cccccc@{}}
\toprule
\textbf{Model} & \textbf{FCAT 1} & \textbf{FCAT 2} & \textbf{FCAT 3} & \textbf{Jama-Coaque 1} & \textbf{Jama-Coaque 2} \\ \midrule
\textbf{RT-DETR}        & 42.9  & 378.0 & 273.5 & 177.1 & 143.0 \\
\textbf{YOLOv8}         & 34.1  & 297.6 & 213.1 & 138.9 & 112.0 \\
\textbf{YOLOv9}         & 34.7  & 297.2 & 216.4 & 140.4 & 113.8 \\
\textbf{YOLOv10}        & 31.3  & 273.8 & 197.5 & 134.3 & 102.3 \\
\midrule
\textbf{SAM}            & 175.9 & 1454.9& 1219.5& 496.6 & 590.4 \\
\textbf{SAM 2}          & 150.4 & 1265.4& 1052.8& 445.7 & 509.1 \\
\textbf{Mobile SAM}     & 107.2 & 835.8 & 654.0 & 341.7 & 334.2 \\
\bottomrule
\end{tabular}%
}
\end{table}

\begin{figure}[t]
    \centering
    \includegraphics[width=0.7\linewidth]{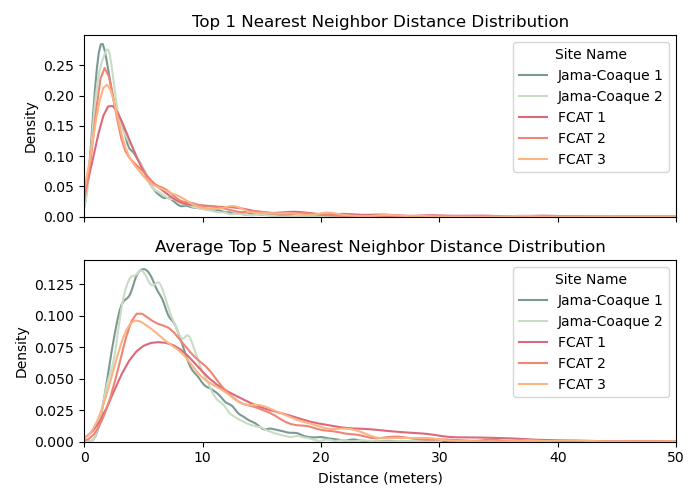}
    \caption{Distribution of Distances to Nearest Neighbors Among Detected Palm Centers Across Sites.}
    \label{fig:cluster}
\end{figure}

\subsubsection{Counting}

Table \ref{tab:effectiveness} evaluates the counting accuracy of detection models across five sites, assessing the impact of training data quantity. Counting accuracy is determined by the proximity of detected centers to labeled centers, with success defined as a center detected within a 5-meter radius of the labeled center. FCAT 1 and FCAT 2, marked with $^\star$, were used for training image patches and exhibit excellent performance at these sites and strong generalizability at FCAT 3, demonstrating robust model adaptability within the same ecological reserve. However, performance varies at different reserves, with RT-DETR consistently excelling at Jama-Coaque sites. Notably, the minimal variance between training with 10\% and 90\% of the data suggests that more extensive training improves object localization precision rather than overall detection efficiency.

Additionally, Figure \ref{fig:kde} quantifies the palm localization performance across five sites by utilizing estimated density and cumulative distributions of distance shifts. These shifts often occur when bounding box centers do not align with actual palm centers, typically when only part of a palm is visible, causing the bounding box center to skew towards visible leaves. FCAT sites have significantly high performance, with mean distances under 0.90 meter and medians under 0.77 meters. In contrast, Jama-Coaque sites, which were not included in the training set, show slightly larger distance shifts, with means under 1.30 meters and medians less than 1.06 meters, yet still within the acceptable range under 5 meters. 

Furthermore, FCAT sites display lower variability in distance shifts, with standard deviations ranging from 0.51 to 0.60 meters, compared to 0.78 to 1.02 meters for Jama-Coaque. This indicates more consistent palm localization in FCAT regions, possibly due to more homogeneous environmental conditions or better palm visibility in the dataset. The slightly larger variance observed in Jama-Coaque reflects the model's greater difficulty in adapting to variability in forest composition and palm species. Nonetheless, the model demonstrates robust adaptability across diverse environments.

\subsubsection{Real-time Processing}
The latency data presented in Table \ref{tab:latency} for PalmDSNet across five sites substantiates the framework's capability for real-time processing, which is essential for immediate ecological assessments. In this context, detection (first 4 rows) refers to the process of identifying palms and saving all relevant information into a CSV file, while segmentation (last 3 rows) involves generating and saving the combined segmentation mask overlaid on the image. As demonstrated previously, our models maintain a high throughput, processing over 40 frames per second on an RTX 3090, which underscores their potential for real-time applications and suitability for on-board UAV deployment. Detection times vary from 31.3 to 378 seconds depending on the area size processed, and segmentation times, guided by palm counts and RT-DETR results, range from 107.2 to 1454.9 seconds. Overall, the processing times for detection, segmentation, and counting span from about two and a half minutes to half an hour per site, confirming the framework's practicality for real-time use in varied environments.

\begin{figure*}[tb]
  \centering
  \begin{subfigure}{0.195\textwidth}
    \includegraphics[width=\textwidth]{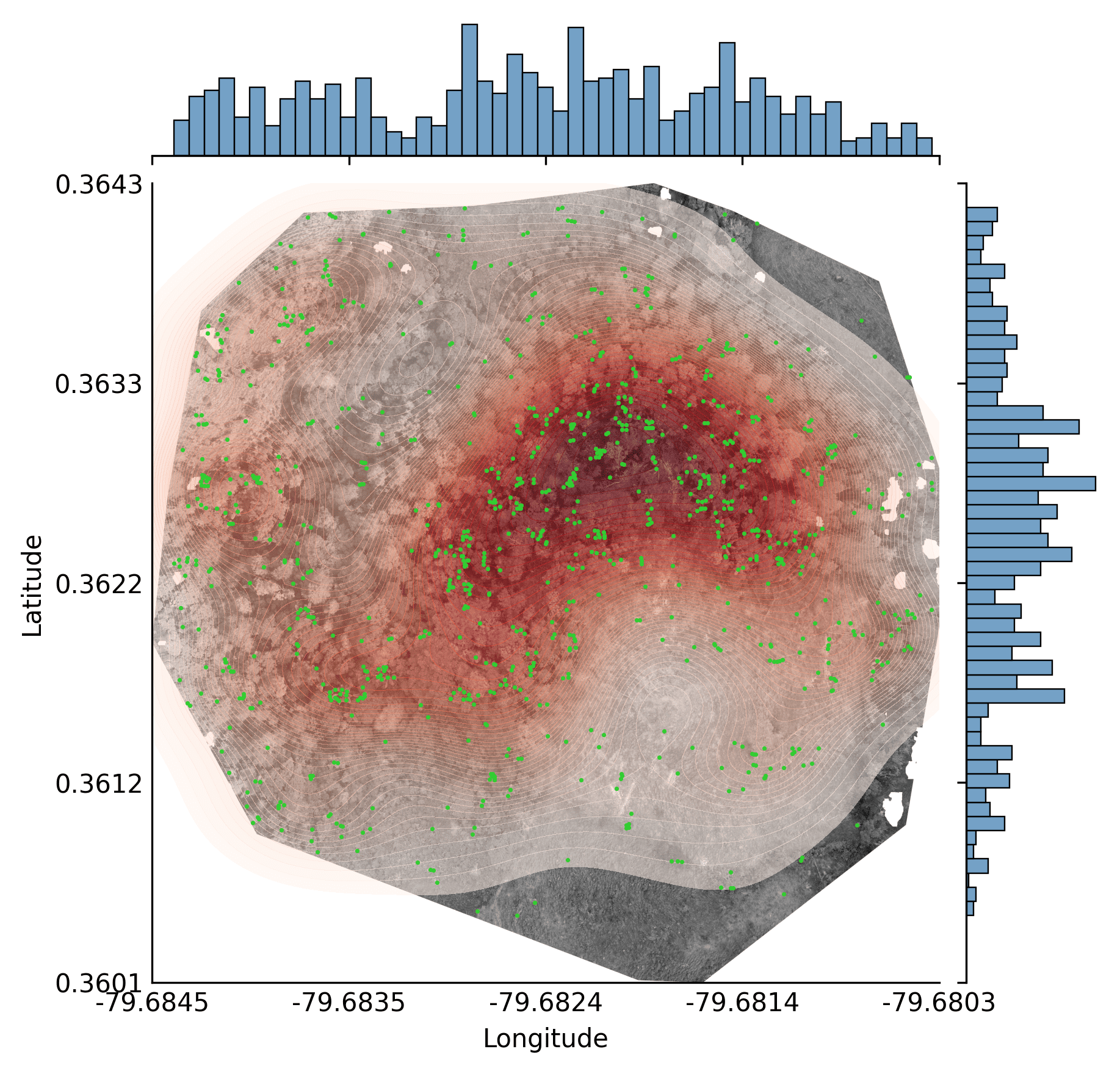}
    \caption{FCAT 1}
  \end{subfigure}%
  \hfill
  \begin{subfigure}{0.195\textwidth}
    \includegraphics[width=\textwidth]{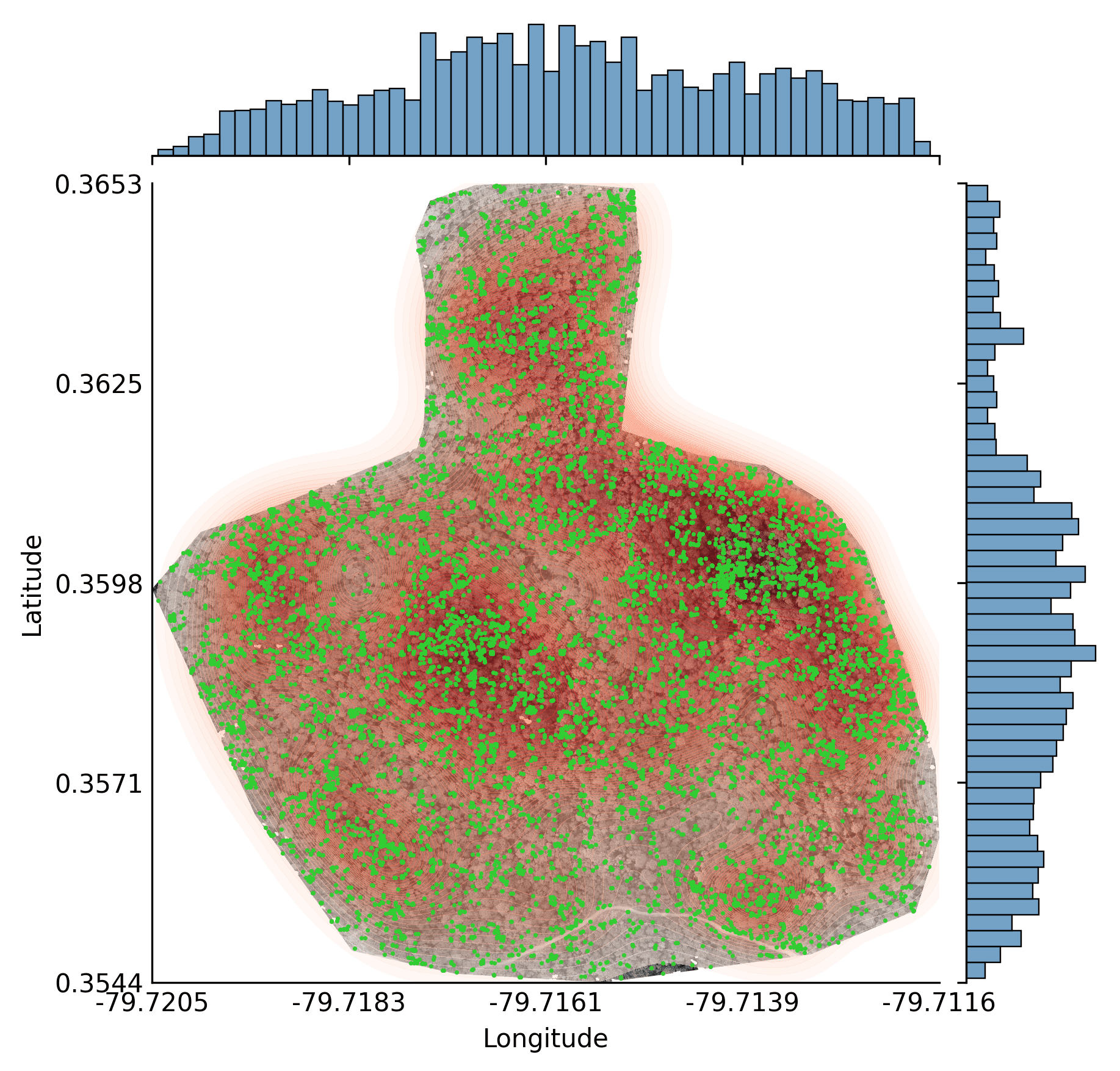}
    \caption{FCAT 2}
  \end{subfigure}%
  \hfill
  \begin{subfigure}{0.195\textwidth}
    \includegraphics[width=\textwidth]{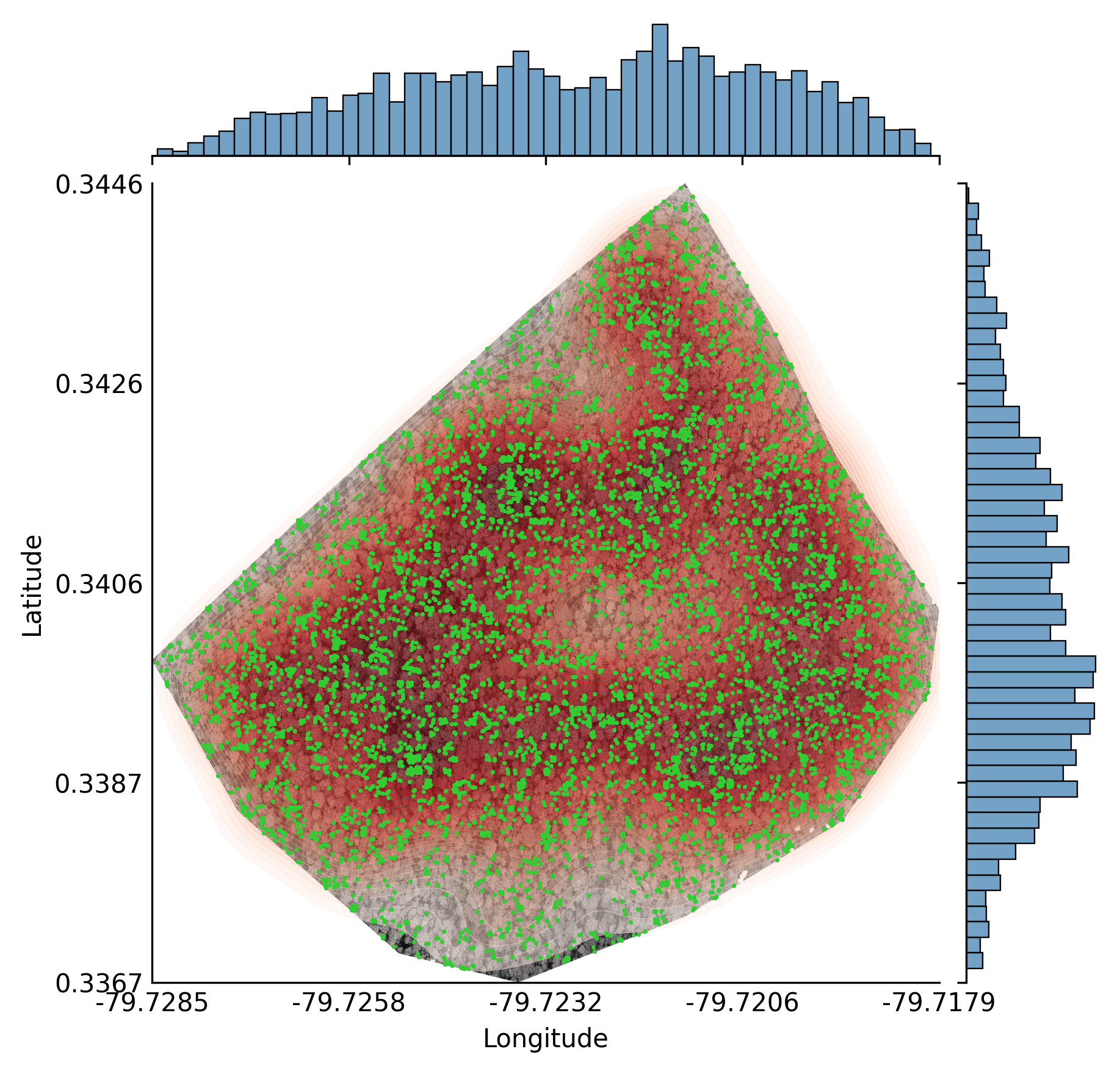}
    \caption{FCAT 3}
  \end{subfigure}%
  \hfill
  \begin{subfigure}{0.195\textwidth}
    \includegraphics[width=\textwidth]{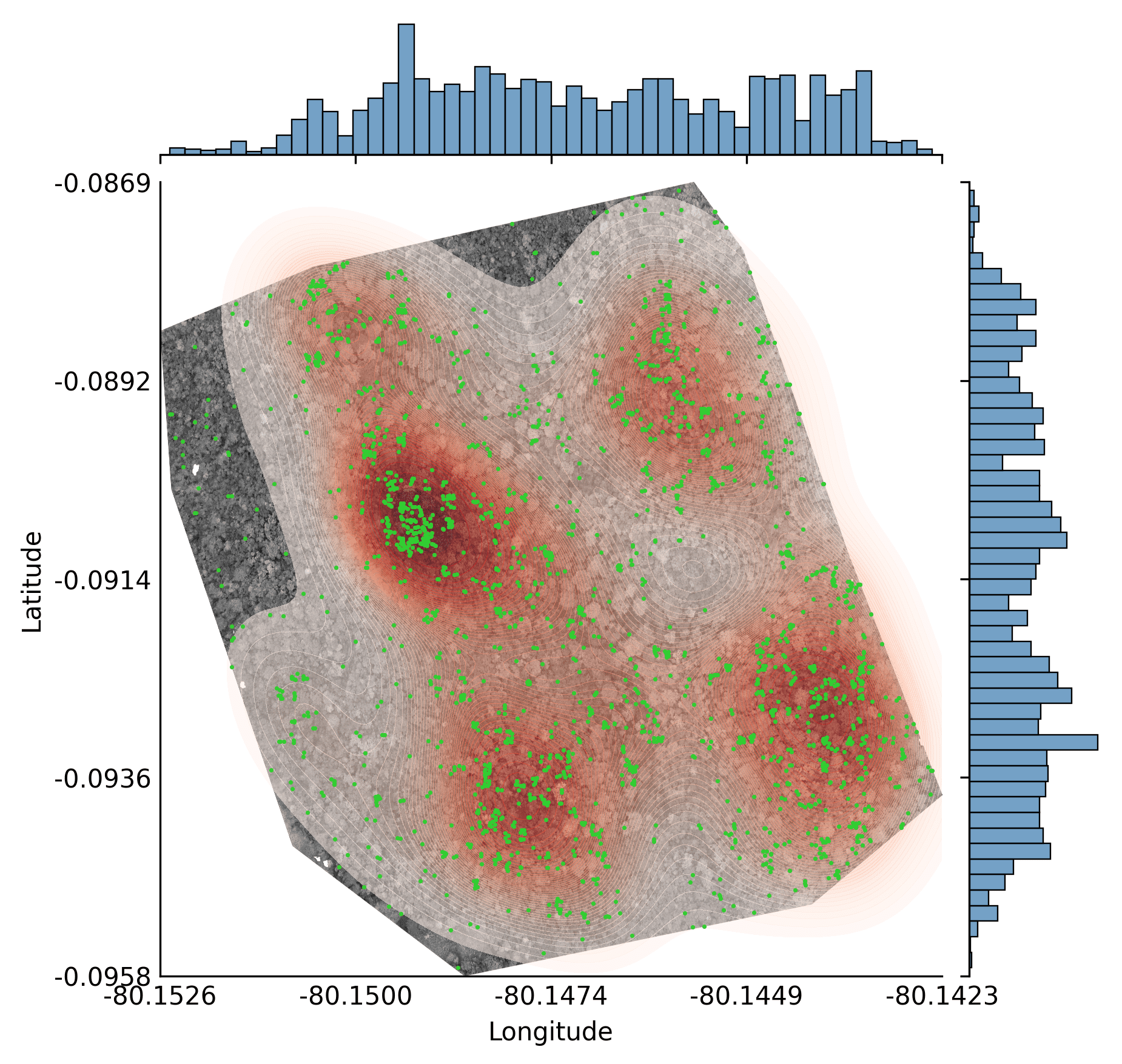}
    \caption{Jama-Coaque 1}
    \label{fig:sub-jama1}
  \end{subfigure}%
  \hfill
  \begin{subfigure}{0.195\textwidth}
    \includegraphics[width=\textwidth]{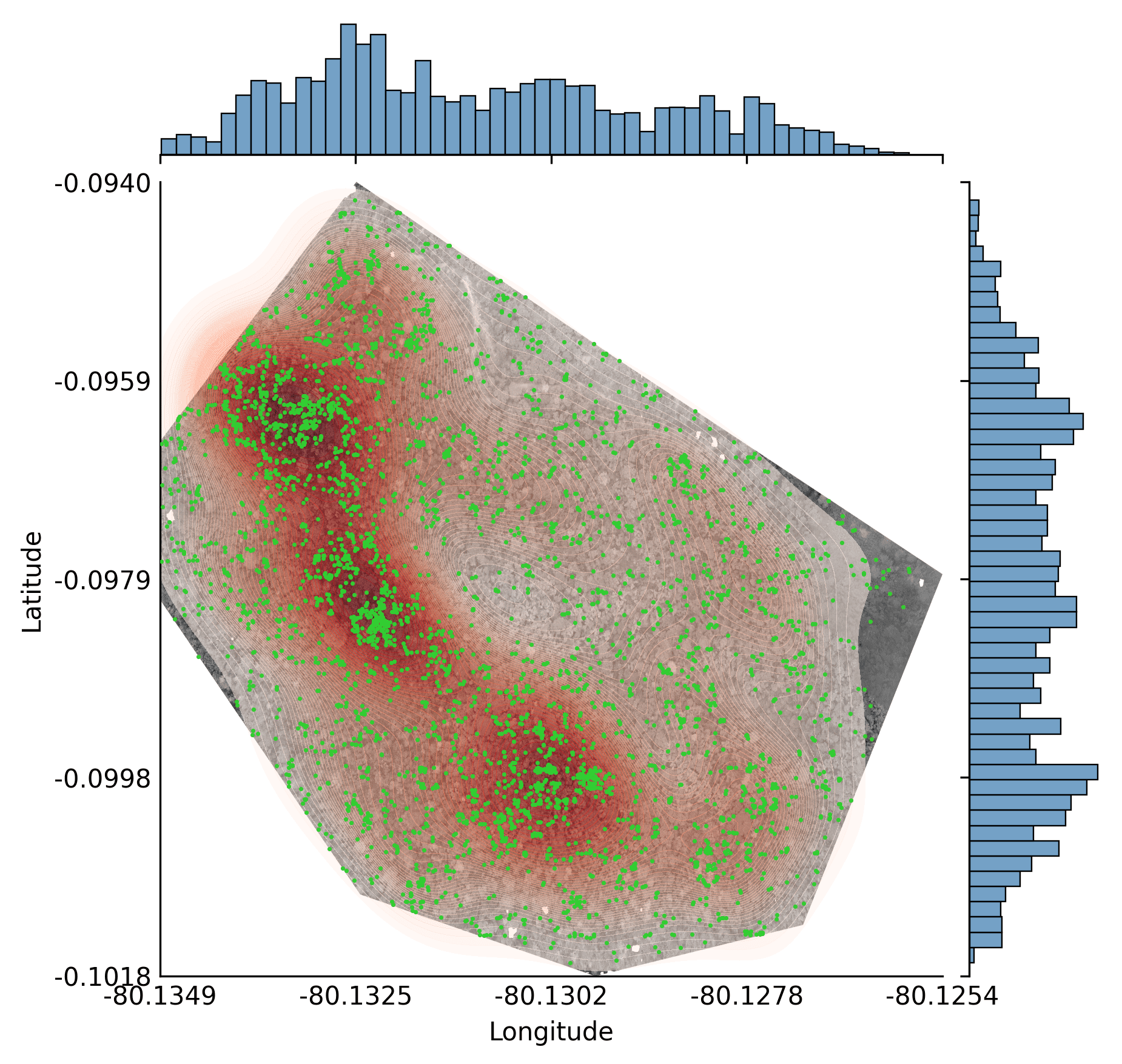}
    \caption{Jama-Coaque 2}
  \end{subfigure}
  
  \vspace{0.5cm} 

  \begin{subfigure}{0.195\textwidth}
    \includegraphics[width=\textwidth]{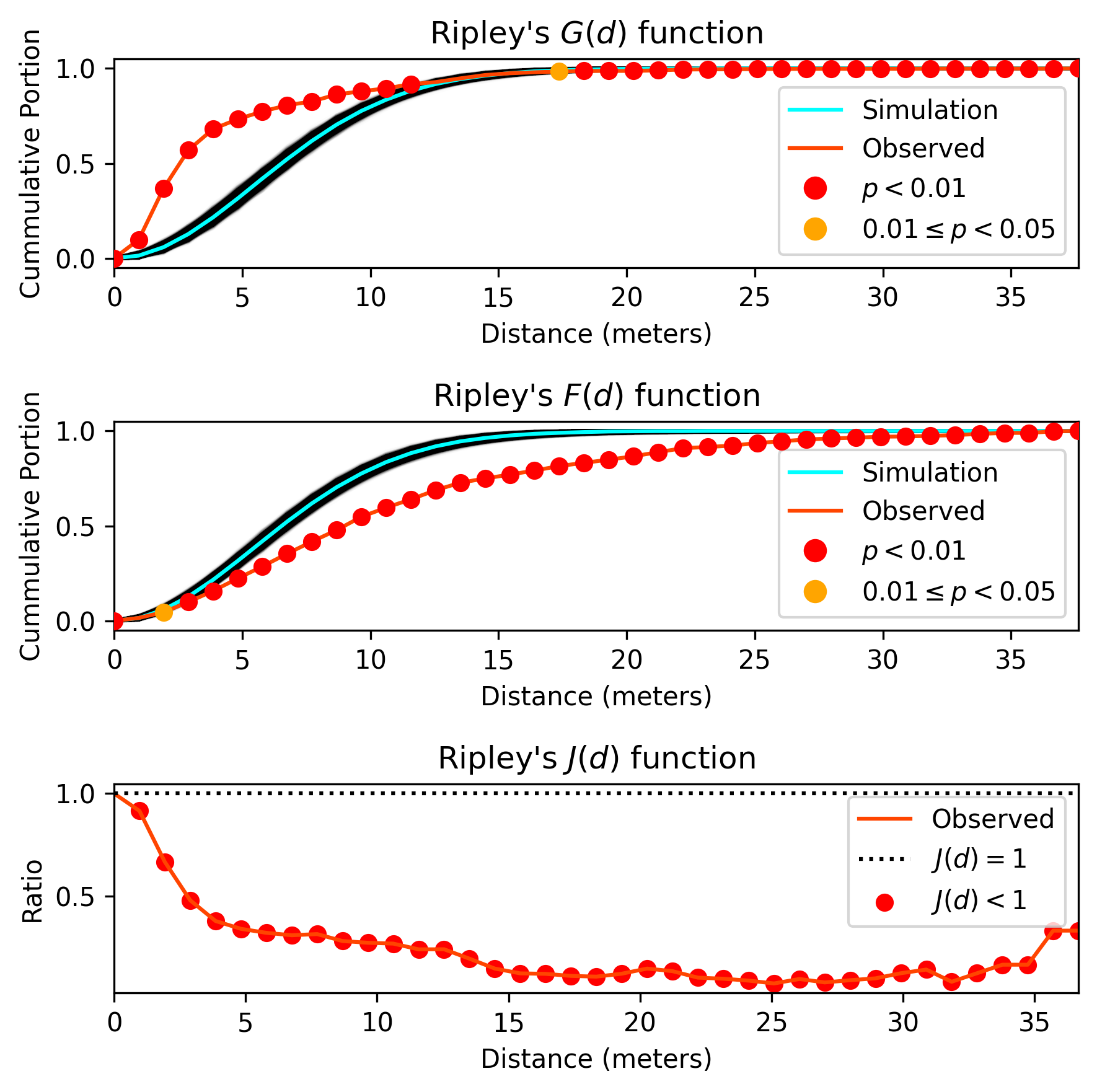}
    \caption{FCAT 1}
  \end{subfigure}%
  \hfill
  \begin{subfigure}{0.195\textwidth}
    \includegraphics[width=\textwidth]{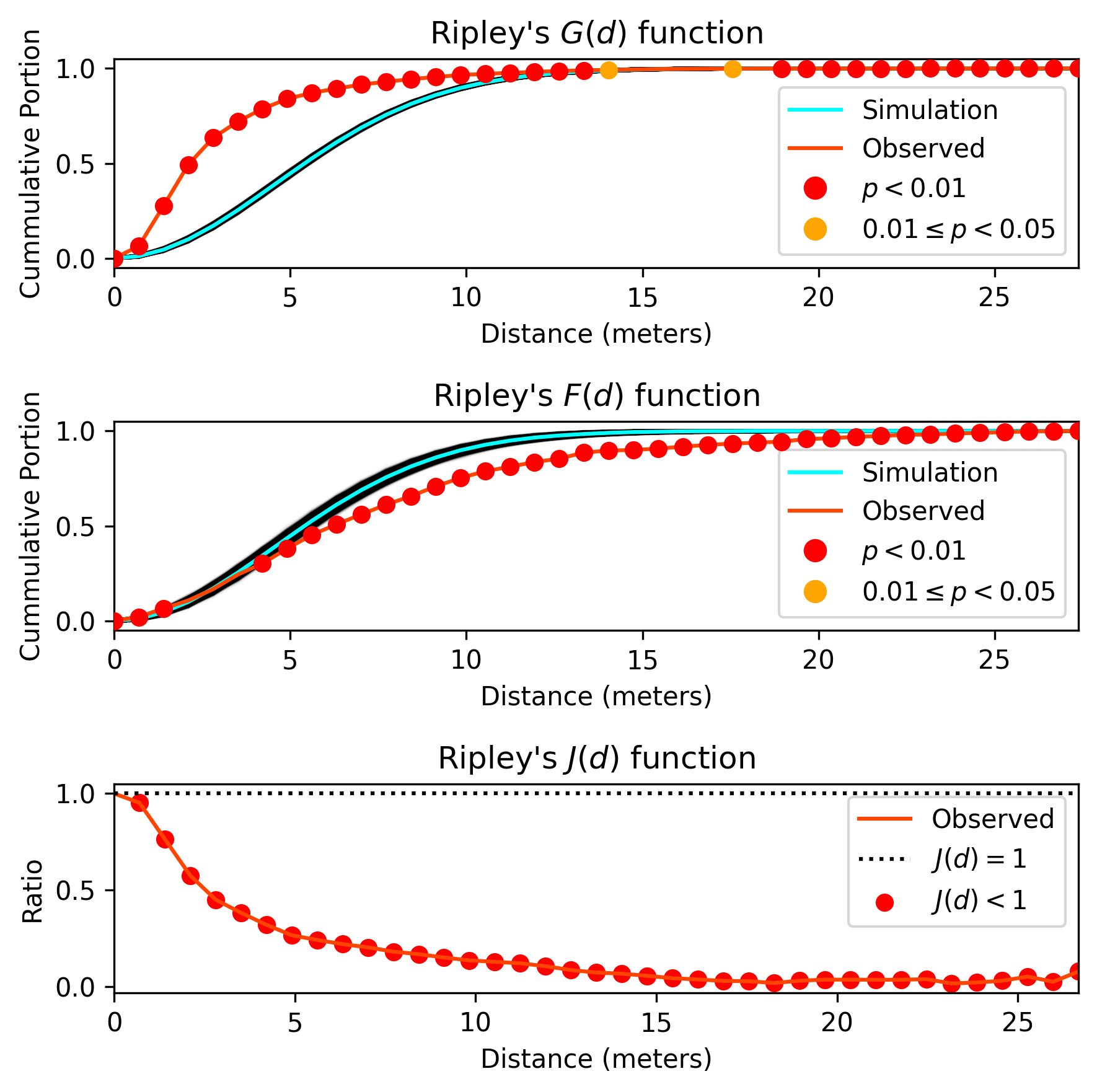}
    \caption{FCAT 2}
  \end{subfigure}%
  \hfill
  \begin{subfigure}{0.195\textwidth}
    \includegraphics[width=\textwidth]{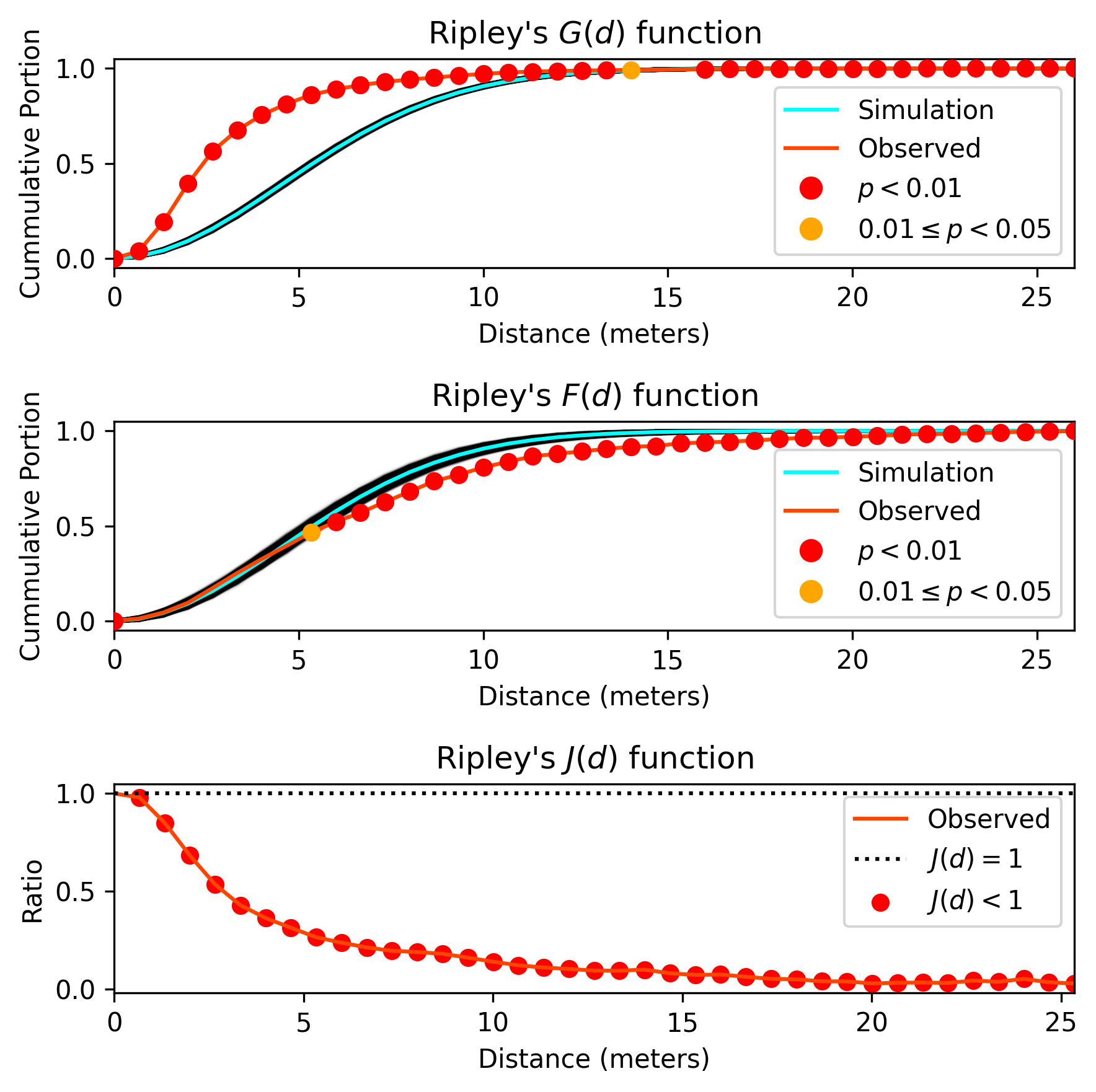}
    \caption{FCAT 3}
  \end{subfigure}%
  \hfill
  \begin{subfigure}{0.195\textwidth}
    \includegraphics[width=\textwidth]{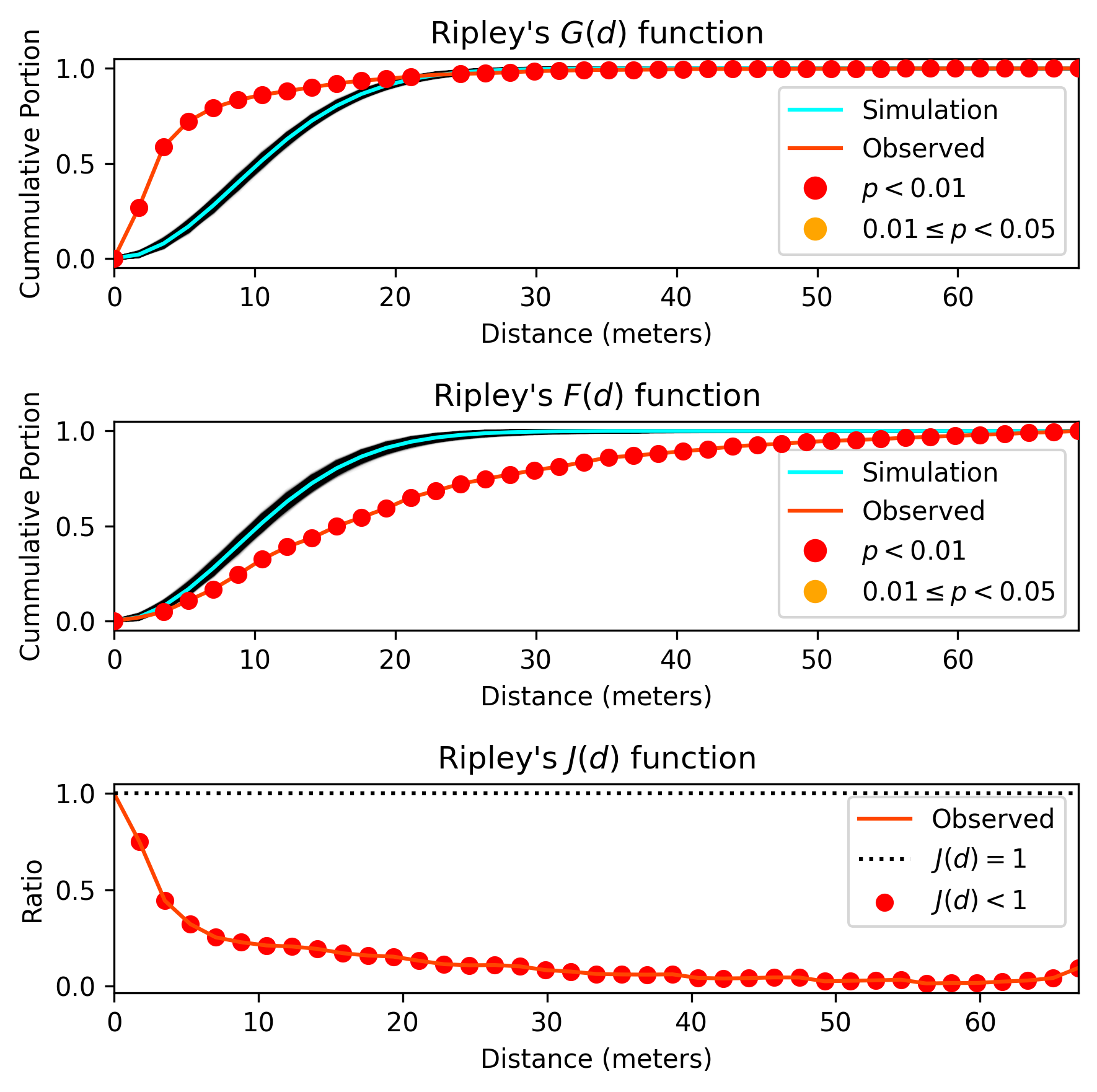}
    \caption{Jama-Coaque 1}
  \end{subfigure}%
  \hfill
  \begin{subfigure}{0.195\textwidth}
    \includegraphics[width=\textwidth]{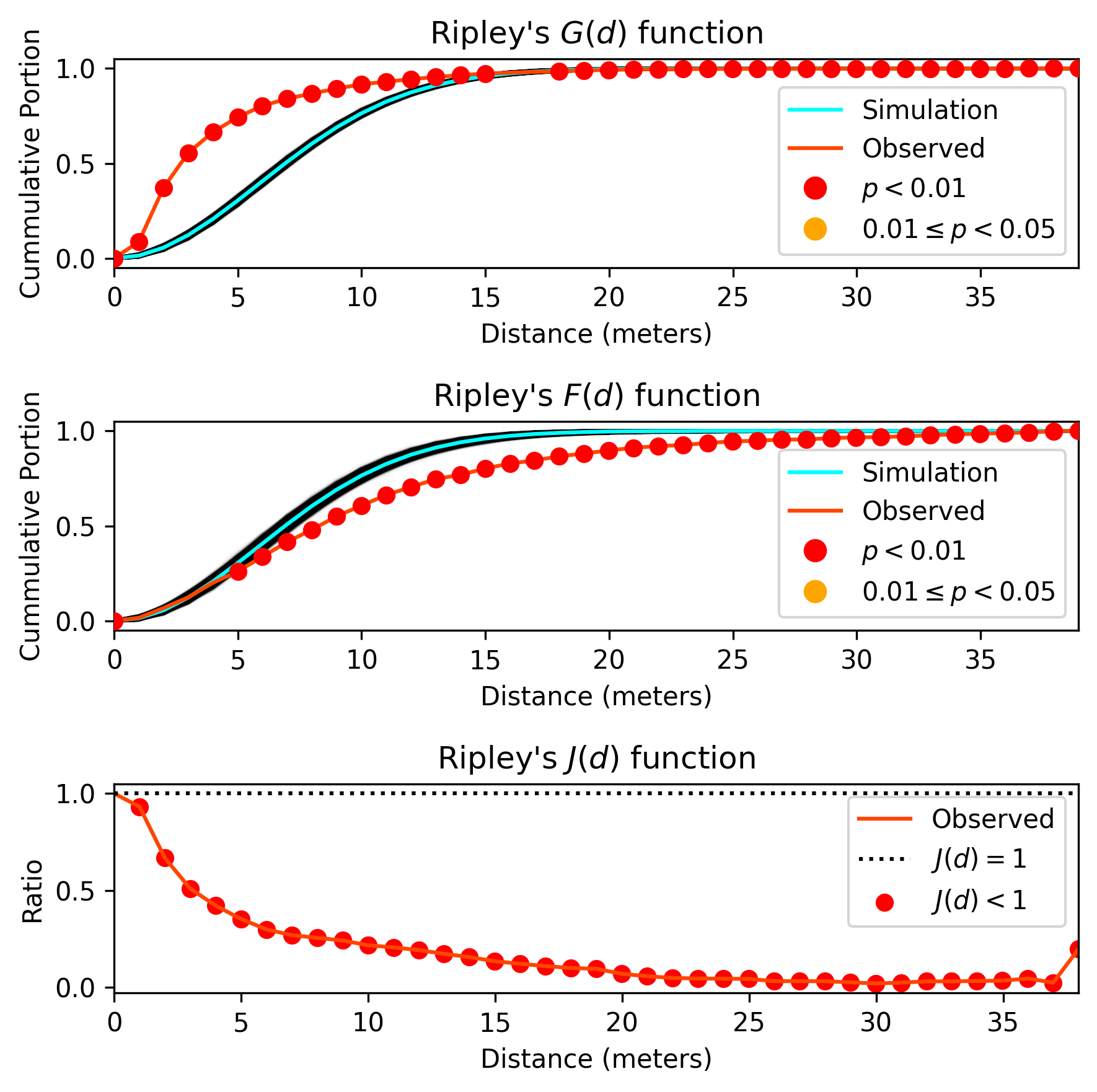}
    \caption{Jama-Coaque 2}
  \end{subfigure}

  \caption{Visualization and Randomness Analysis of Palm Spatial Distribution. The first row (a-e) presents kernel density estimates with histograms illustrating palm distribution across regions. Green dots indicate detected palm locations, with darker KDE regions showing higher clustering. The second row (f-j) compares Ripley's functions of the detected palms to those generated by a Poisson process, highlighting deviations from randomness. Blue-shaded areas represent the 95\% confidence intervals with the black line indicating the mean for a Poisson process, while the red lines represent Ripley's functions of the observed data. Red and orange points denote regions with $p$-values below 1\% and 5\%, respectively, indicating significant differences from randomness. These indices quantify palm clustering or dispersion relative to a random distribution.}
  \label{fig:spatial_analysis}
\end{figure*}

\subsection{Spatial Distribution Analysis}

The analysis of spatial distribution in palm populations provides critical insights into their ecological dynamics and interspecies relationships. By examining spatial patterns, we can better understand species coexistence and competition, which are essential for guiding sustainable forest management and utilization strategies~\cite{tao2021mapping, ben2021spatial}. This section explores various aspects of palm spatial distribution, including nearest neighbor analysis, distribution randomness, and simulation of spatial point patterns.

\begin{table}[tb]
  \centering
  \caption{Parameters $p^*$ and $\sigma^*$ for different sites.}
  \resizebox{0.6\linewidth}{!}{%
    \begin{tabular}{cccccc}
      \toprule
      \textbf{Parameters} & \textbf{FCAT 1} & \textbf{FCAT 2} & \textbf{FCAT 3} & \textbf{Jama-Coaque 1} & \textbf{Jama-Coaque 2} \\
      \midrule
      $p^*$ & 0.49 & 0.52 & 0.46 & 0.64 & 0.51 \\
      $\sigma^*$ & 50 & 70 & 70 & 80 & 60 \\
      \bottomrule
    \end{tabular}
  }
  \label{tab:pointpats}
\end{table}

\begin{figure*}[tb]
  \centering
  \begin{subfigure}{0.85\linewidth}
    \includegraphics[width=\textwidth]{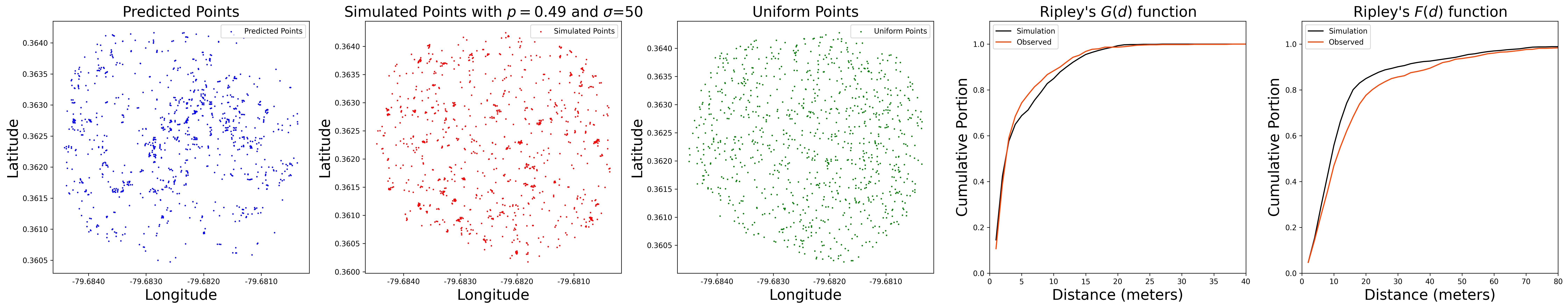}
    \caption{FCAT 1}
  \end{subfigure}%
  
  \begin{subfigure}{0.85\linewidth}
    \includegraphics[width=\textwidth]{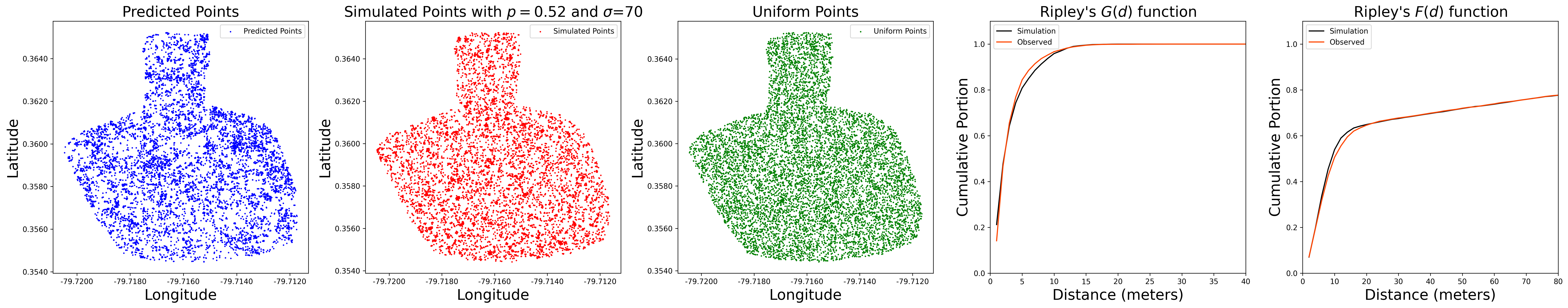}
    \caption{FCAT 2}
  \end{subfigure}%
  
  \begin{subfigure}{0.85\linewidth}
    \includegraphics[width=\textwidth]{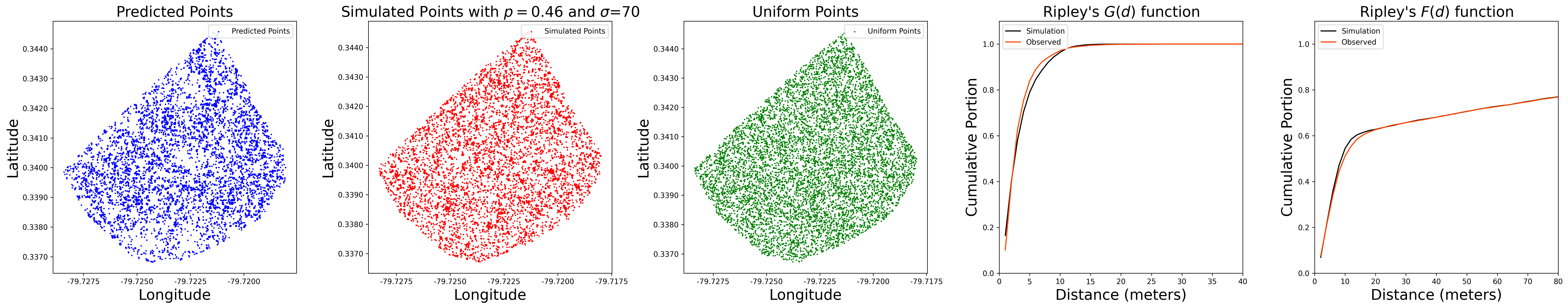}
    \caption{FCAT 3}
  \end{subfigure}%

  \begin{subfigure}{0.85\linewidth}
    \includegraphics[width=\textwidth]{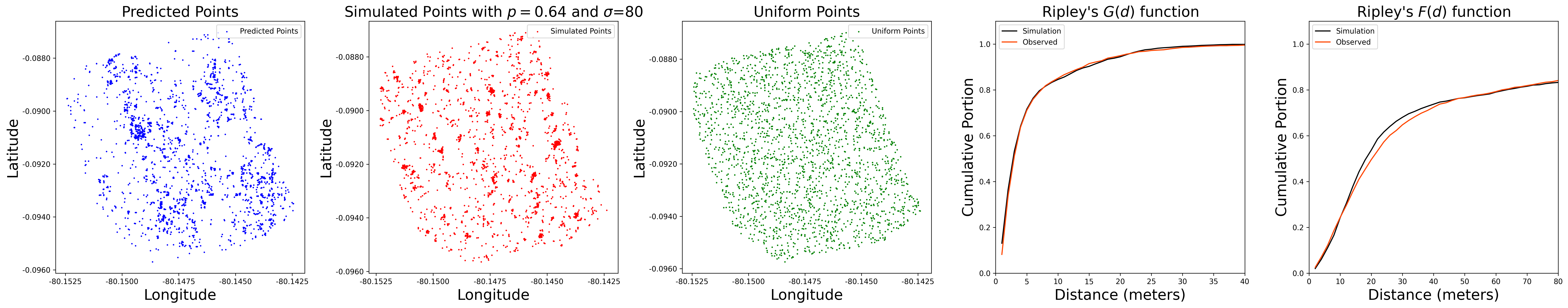}
    \caption{Jama-Coaque 1}
  \end{subfigure}%

  \begin{subfigure}{0.85\linewidth}
    \includegraphics[width=\textwidth]{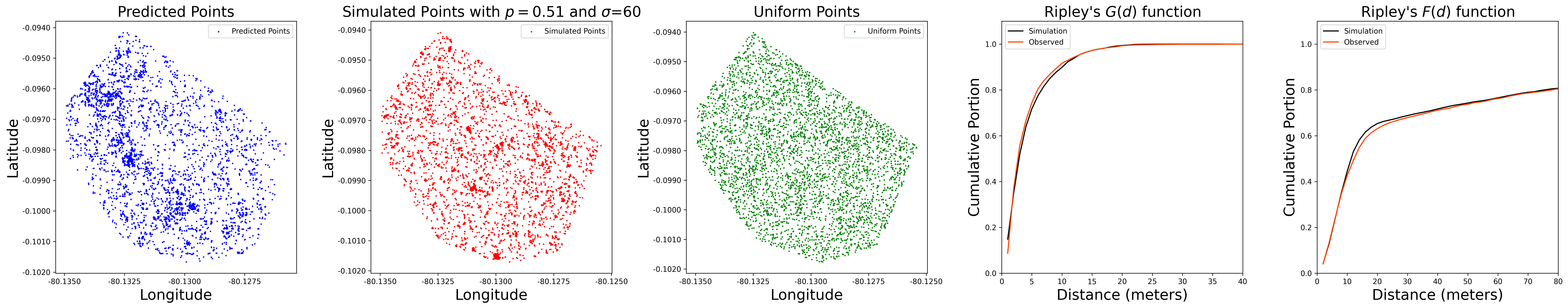}
    \caption{Jama-Coaque 2}
  \end{subfigure}%
  
  \caption{Comparison of Simulated and Predicted Point Patterns. The first column presents the predicted spatial distribution of detected palms by PalmDSNet, the second column shows the simulated distribution, and the third column depicts random points with a uniform distribution. The fourth and fifth columns compare Ripley's $G$ and $F$ functions, respectively, between the simulated and predicted distributions.}
  \label{fig:pointpats}
\end{figure*}

\subsubsection{Analysis of Spatial Nearest Neighbors}

Figure \ref{fig:cluster} depicts the distribution of distances from each detected palm to its nearest neighbors to explore their spatial arrangement. The distribution of nearest neighbor distance reveals a compact clustering of palms, indicating that palms often grow in close proximity, which suggests potential biological interactions or shared habitat preferences. The average top five nearest neighbors' distance distribution shows that while palms generally form tight clusters, they also participate in larger, slightly more dispersed groups. Notably, this pattern is more distinct in Jama-Coaque than in FCAT, which reflects diverse ecological strategies and community structuring across different reserves, with the contribution of endogenous demographic processes and environmental factors being a promising area for future research.

\subsubsection{Analysis of Distribution Randomness}

The primary objective of this analysis is to determine whether the spatial distribution of palms across various sites deviates from a random pattern. By comparing the observed palm distributions with those generated by a Poisson point process, we aim to identify underlying ecological patterns that may indicate clustering or dispersion within these environments.

Figure~\ref{fig:spatial_analysis} illustrates the spatial distribution of palms across the five study sites. These regions display varying levels of palm density, with certain areas exhibiting notably higher concentrations than others. The visualizations in the first row reveal that the palm distributions are not random; clusters and dispersed regions are clearly visible, suggesting non-random spatial arrangements. The second row of Figure~\ref{fig:spatial_analysis} compares the observed point patterns of palm distributions with those generated by a Poisson point process using Ripley's functions. For the $G$ and $F$ plots, the central black line represents the expected pattern if palms were randomly scattered throughout the forest. The blue-shaded area around this line indicates the 95\% confidence interval within which random data points would typically fall, with different confidence levels ($1-p$) for the sampled points being random, depicted with varying colors for $p < 0.01$ and $p < 0.05$. The $J$ function plot includes a dashed line at $J(d) = 1$, with values below this line colored red to indicate clustering.

For the $G$ function, the observed line ascends more steeply than the random case, indicating a higher degree of clustering among the palms, with palms closer to their nearest neighbors than they would be in a completely random distribution. Conversely, in the $F$ function, the observed line falls below the random case, implying that random points must travel a greater distance to find a nearby palm, indicating larger gaps or empty spaces in the actual palm distribution compared to a random arrangement. The $J$ function, with values below 1, further supports these observations by indicating that the palms are more clustered than would be expected under randomness. Together, these Ripley's functions reveal significant deviations from randomness and suggest complex ecological interactions shaping palm distributions. The observed pattern's clustering is more prominent than randomness yet less extreme than one would expect with a Student's $t$ distribution. This sets the stage for future work to explore the underlying ecological drivers of the distribution.

\subsubsection{Simulation of Spatial Point Pattern}

Given the statistically significant evidence that the distribution of palms is far from random, as anticipated due to the influence of various factors such as habitat suitability, as well as human and animal activities, we seek to employ a statistical yet straightforward model that can simulate and explain the spatial point patterns of palm distribution. To this end, we utilize the proposed Poisson-Gaussian palm reproduction model, which integrates the inherent stochastic nature of palm reproduction with spatially dependent factors, thereby generating distributions that can be compared with those predicted by PalmDSNet. 

To measure clustering and dispersion in the simulated palm distributions, we use Ripley's $G$ and $F$ functions, which quantify spatial patterns, with $G$ indicating clustering and $F$ showing dispersion. By comparing these functions from our simulations to those predicted by PalmDSNet, we aim to align the simulated distributions with the predicted ones. Ecologically, the parameter $p$ indicates the likelihood of global randomness in palm distribution, with higher values denoting greater randomness and lower values indicating local clustering with a Gaussian distribution. The parameter $\sigma$ represents the local range of palm reproduction, reflecting the degree of clustering around individual palms. 

Table~\ref{tab:pointpats} lists the grid search results of the $p^*$ and $\sigma^*$ pairs for the five regions. Across the five sites, $\sigma^*$ consistently ranges from 50 to 80 ($\sim$2$-$4 m), and $p^*$ values for four of the sites range from 0.46 to 0.52, both of which are quite stable. Jama-Coaque 1 is an exception, with a value of $p^*$ of 0.64, indicating more dispersed clusters, as also reflected by the slightly larger $\sigma^*$ for this site. This greater dispersion is evident in the estimated density shown in Figure~\ref{fig:sub-jama1}. The robustness of $p^*$ and $\sigma^*$ across different forests demonstrates the model's capability to capture the common point patterns of palm distributions, even with only two parameters. This simplicity helps in understanding the spread of palm species in forests under various scenarios.

Figure~\ref{fig:pointpats} compares our simulated point patterns to those predicted across five sites. The first column presents the predicted spatial distribution of detected palms by PalmDSNet, the second column shows the simulated distribution, and the third column depicts random points with a uniform distribution. The fourth and fifth columns compare Ripley's $G$ and $F$ functions, respectively, between the simulated and predicted distributions. This last comparison shows an excellent fit for all sites. With the exception of FCAT 1, the simulated point patterns in the second column are visually more homogeneous than the predicted ones. This is easily explained by the fact that the distribution parameters were estimated globally. This observation should lead us to perform slightly more localized parameter estimation in the future.

\section{Conclusion}
\label{sec:conclude}

In summary, our work with the PalmDSNet framework has proven to be effective in detecting, segmenting, and counting palms across diverse forest environments. We found that increasing the number of labels enhances the precision of bounding boxes, though it has a minimal impact on counting accuracy. The framework is also capable of real-time processing, even under resource-constrained conditions, including on-board UAV processing and integration into autonomous flight control systems. Additionally, our examination of palm distribution across two different forest reserves reveals that palms often form tightly-knit clusters, yet their spatial arrangements can differ significantly between forest ecosystems. The Poisson-Gaussian reproduction model, which employs a brute-force method to identify the best-fitting parameters that align with observed point patterns using Ripley's $G$ and $F$ functions, has also shown satisfying performance. This model reliably captures the spatial patterns of palm distribution across various forest reserves, with consistent parameter values observed in most sites. By quantifying the level of clustering and its deviation from randomness, the Poisson-Gaussian reproduction model enhances our understanding of spatial processes of palms, which is crucial for environmental monitoring, conservation efforts, management strategies, economic planning, and supply chain optimization. Together, these two models provide an end-to-end solution, from image analysis to addressing ecological challenges.

In future work, we intend to expand our dataset through a semi-manual approach, wherein our trained model will assist in the labeling process. This expansion will include the annotation of additional sites in Ecuador, where data collection is feasible, enabling us to conduct species-level mapping to explore how spatial patterns evolve from the wettest forests in the Chocó to the limit of the tropical dry forest at the Sechura Desert, as well as a 1000 km$^2$ area in Amazonian forest near Iquitos, Peru. Besides using orthomosaics, we want to compare whether multiple views from raw UAV images, without processing, provide deeper insights into the canopy and improve detection of partially occluded palms, potentially revealing more hidden individuals. The established framework can also be readily adapted to address other conservation challenges~\cite{camalan2022change}, such as species-level identification of canopy trees,  as well as the detection and segmentation of agriculture and forestry,  and water bodies -- the latter a task of significant importance for monitoring illegal mining activities and other environmental concerns~\cite{cui2022semi, camalan2022detecting}. Furthermore, we will consider integrating more complex statistical models to refine the alignment of spatial patterns, including quantifying the importance of endogenous demographic processes as well as environmental predictors, and historical and current human effects on palm populations, particularly after incorporating species-level classifications.

\printbibliography

@article{sutherland2013identification,
  title={Identification of 100 fundamental ecological questions},
  author={Sutherland, William J and Freckleton, Robert P and Godfray, H Charles J and Beissinger, Steven R and Benton, Tim and Cameron, Duncan D and others},
  journal={Journal of ecology},
  volume={101},
  number={1},
  pages={58--67},
  year={2013},
  publisher={Wiley Online Library}
}

@article{wagner2020regional,
  title={Regional mapping and spatial distribution analysis of canopy palms in an amazon forest using deep learning and VHR images},
  author={Wagner, Fabien H and Dalagnol, Ricardo and Tagle Casapia, Ximena and Streher, Annia S and Phillips, Oliver L and Gloor, Emanuel and Arag{\~a}o, Luiz EOC},
  journal={Remote Sensing},
  volume={12},
  number={14},
  pages={2225},
  year={2020},
  publisher={MDPI}
}

@inproceedings{cui2024palmprobnet,
  title={PalmProbNet: A Probabilistic Approach to Understanding Palm Distributions in Ecuadorian Tropical Forest via Transfer Learning},
  author={Cui, Kangning and Shao, Zishan and Larsen, Gregory and Pauca, Victor and Alqahtani, Sarra and Segurado, David and others},
  booktitle={Proceedings of the 2024 ACM Southeast Conference},
  pages={272--277},
  year={2024}
}

@article{malhi2014tropical,
  title={Tropical forests in the Anthropocene},
  author={Malhi, Yadvinder and Gardner, Toby A and Goldsmith, Gregory R and Silman, Miles R and Zelazowski, Przemyslaw},
  journal={Annual Review of Environment and Resources},
  volume={39},
  pages={125--159},
  year={2014},
  publisher={Annual Reviews}
}

@article{pitman2014distribution,
  title={Distribution and abundance of tree species in swamp forests of Amazonian Ecuador},
  author={Pitman, Nigel CA and Andino, Juan Ernesto Guevara and Aulestia, Milton and Cer{\'o}n, Carlos E and Neill, David A and Palacios, Walter and others},
  journal={Ecography},
  volume={37},
  number={9},
  pages={902--915},
  year={2014},
  publisher={Wiley Online Library}
}

@article{zhang2023aerial,
  title={Aerial orthoimage generation for UAV remote sensing},
  author={Zhang, Jiguang and Xu, Shilin and Zhao, Yong and Sun, Jiaxi and Xu, Shibiao and Zhang, Xiaopeng},
  journal={Information Fusion},
  volume={89},
  pages={91--120},
  year={2023},
  publisher={Elsevier}
}

@inproceedings{cui2021unsupervised,
  title={Unsupervised classification of AVIRIS-NG hyperspectral images},
  author={Cui, Kangning and Plemmons, Robert J},
  booktitle={2021 11th Workshop on Hyperspectral Imaging and Signal Processing: Evolution in Remote Sensing (WHISPERS)},
  pages={1--5},
  year={2021},
  organization={IEEE}
}

@article{freudenberg2019large,
  title={Large scale palm tree detection in high resolution satellite images using U-Net},
  author={Freudenberg, Maximilian and N{\"o}lke, Nils and Agostini, Alejandro and Urban, Kira and W{\"o}rg{\"o}tter, Florentin and Kleinn, Christoph},
  journal={Remote Sensing},
  volume={11},
  number={3},
  pages={312},
  year={2019},
  publisher={MDPI}
}

@article{gibril2021deep,
  title={Deep convolutional neural network for large-scale date palm tree mapping from UAV-based images},
  author={Gibril, Mohamed Barakat A and Shafri, Helmi Zulhaidi Mohd and Shanableh, Abdallah and Al-Ruzouq, Rami and Wayayok, Aimrun and Hashim, Shaiful Jahari},
  journal={Remote Sensing},
  volume={13},
  number={14},
  pages={2787},
  year={2021},
  publisher={MDPI}
}

@article{li2016deep,
  title={Deep Learning Based Oil Palm Tree Detection and Counting for High-Resolution Remote Sensing Images},
  author={Li, Weijia and Fu, Haohuan and Yu, Le and Cracknell, Arthur},
  journal={Remote Sensing},
  volume={9},
  number={1},
  pages={22},
  year={2016},
  publisher={MDPI}
}

@article{qin2021ag,
  title={Ag-YOLO: A real-time low-cost detector for precise spraying with case study of palms},
  author={Qin, Zhenwang and Wang, Wensheng and Dammer, Karl-Heinz and Guo, Leifeng and Cao, Zhen},
  journal={Frontiers in Plant Science},
  volume={12},
  pages={753603},
  year={2021},
  publisher={Frontiers Media SA}
}

@article{jintasuttisak2022deep,
  title={Deep neural network based date palm tree detection in drone imagery},
  author={Jintasuttisak, Thani and Edirisinghe, Eran and Elbattay, Ali},
  journal={Computers and Electronics in Agriculture},
  volume={192},
  pages={106560},
  year={2022},
  publisher={Elsevier}
}

@misc{mobile_sam,
  title={Faster Segment Anything: Towards Lightweight SAM for Mobile Applications},
  author={Zhang, Chaoning and Han, Dongshen and Qiao, Yu and Kim, Jung Uk and Bae, Sung-Ho and Lee, Seungkyu and Hong, Choong Seon},
  year={2023},
  eprint={2306.14289},
  archivePrefix={arXiv},
  note={arXiv preprint arXiv:2306.14289},
}

@article{zou2023object,
  title={Object detection in 20 years: A survey},
  author={Zou, Zhengxia and Chen, Keyan and Shi, Zhenwei and Guo, Yuhong and Ye, Jieping},
  journal={Proceedings of the IEEE},
  volume={111},
  number={3},
  pages={257--276},
  year={2023},
  publisher={IEEE}
}

@misc{zhao2023detrs,
  title={Detrs beat yolos on real-time object detection},
  author={Zhao, Yian and Lv, Wenyu and Xu, Shangliang and Wei, Jinman and Wang, Guanzhong and Dang, Qingqing and others},
  year={2023},
  eprint={2304.08069},
  archivePrefix={arXiv},
  note={arXiv preprint arXiv:2304.08069},
}

@inproceedings{carion2020end,
  title={End-to-end object detection with transformers},
  author={Carion, Nicolas and Massa, Francisco and Synnaeve, Gabriel and Usunier, Nicolas and Kirillov, Alexander and Zagoruyko, Sergey},
  booktitle={European conference on computer vision},
  pages={213--229},
  year={2020},
  organization={Springer}
}

@inproceedings{kirillov2023segment,
  title={Segment anything},
  author={Kirillov, Alexander and Mintun, Eric and Ravi, Nikhila and Mao, Hanzi and Rolland, Chloe and Gustafson, Laura and others},
  booktitle={Proceedings of the IEEE/CVF International Conference on Computer Vision},
  pages={4015--4026},
  year={2023}
}

@misc{zhang2023comprehensive,
  title={A comprehensive survey on segment anything model for vision and beyond},
  author={Zhang, Chunhui and Liu, Li and Cui, Yawen and Huang, Guanjie and Lin, Weilin and Yang, Yiqian and Hu, Yuehong},
  year={2023},
  eprint={2305.08196},
  archivePrefix={arXiv},
  note={arXiv preprint arXiv:2305.08196},
}

@misc{yolov8,
  author = {Glenn Jocher and Ayush Chaurasia and Jing Qiu},
  title = {Ultralytics YOLOv8},
  version = {8.0.0},
  year = {2023},
  howpublished = {\url{https://github.com/ultralytics/ultralytics}},
  note = {Accessed: 2024-05-10},
  orcid = {0000-0001-5950-6979, 0000-0002-7603-6750, 0000-0003-3783-7069},
  license = {AGPL-3.0}
}

@misc{yolov9,
  title={YOLOv9: Learning What You Want to Learn Using Programmable Gradient Information}, 
  author={Chien-Yao Wang and I-Hau Yeh and Hong-Yuan Mark Liao},
  year={2024},
  eprint={2402.13616},
  archivePrefix={arXiv},
  primaryClass={cs.CV},
  note={arXiv preprint arXiv:2402.13616}
}

@misc{yolov10,
  title={YOLOv10: Real-Time End-to-End Object Detection}, 
  author={Ao Wang and Hui Chen and Lihao Liu and Kai Chen and Zijia Lin and Jungong Han and Guiguang Ding},
  year={2024},
  eprint={2405.14458},
  archivePrefix={arXiv},
  primaryClass={cs.CV},
  note={arXiv preprint arXiv:2405.14458}
}

@article{diwan2023object,
  title={Object detection using YOLO: Challenges, architectural successors, datasets and applications},
  author={Diwan, Tausif and Anirudh, G and Tembhurne, Jitendra V},
  journal={multimedia Tools and Applications},
  volume={82},
  number={6},
  pages={9243--9275},
  year={2023},
  publisher={Springer}
}

@inproceedings{wang2020cspnet,
  title={CSPNet: A new backbone that can enhance learning capability of CNN},
  author={Wang, Chien-Yao and Liao, Hong-Yuan Mark and Wu, Yueh-Hua and Chen, Ping-Yang and Hsieh, Jun-Wei and Yeh, I-Hau},
  booktitle={Proceedings of the IEEE/CVF conference on computer vision and pattern recognition workshops},
  pages={390--391},
  year={2020}
}

@misc{jacobgilpytorchcam,
  title={PyTorch library for CAM methods},
  author={Jacob Gildenblat and contributors},
  year={2021},
  publisher={GitHub},
  howpublished={\url{https://github.com/jacobgil/pytorch-grad-cam}},
}

@article{muscarella2020global,
  title={The global abundance of tree palms},
  author={Muscarella, Robert and Emilio, Thaise and Phillips, Oliver L and Lewis, Simon L and Slik, Ferry and Baker, William J and others},
  journal={Global Ecology and Biogeography},
  volume={29},
  number={9},
  pages={1495--1514},
  year={2020},
  publisher={Wiley Online Library}
}

@article{hidalgo2022sustainable,
  title={Sustainable palm fruit harvesting as a pathway to conserve Amazon peatland forests},
  author={Hidalgo Pizango, C Gabriel and Honorio Coronado, Eur{\'\i}dice N and del {\'A}guila-Pasquel, Jhon and Flores Llampazo, Gerardo and de Jong, Johan and others},
  journal={Nature Sustainability},
  volume={5},
  number={6},
  pages={479--487},
  year={2022},
  publisher={Nature Publishing Group UK London}
}

@article{eiserhardt2011geographical,
  title={Geographical ecology of the palms (Arecaceae): determinants of diversity and distributions across spatial scales},
  author={Eiserhardt, Wolf L and Svenning, Jens-Christian and Kissling, W Daniel and Balslev, Henrik},
  journal={Annals of Botany},
  volume={108},
  number={8},
  pages={1391--1416},
  year={2011},
  publisher={Oxford University Press}
}

@article{zambrana2007diversity,
  title={Diversity of palm uses in the western Amazon},
  author={Zambrana, Narel Y Paniagua and Byg, Anja and Svenning, Jens-Christian and Moraes, Monica and Grandez, Cesar and Balslev, Henrik},
  journal={Biodiversity and Conservation},
  volume={16},
  pages={2771--2787},
  year={2007},
  publisher={Springer}
}

@article{tagle2019identifying,
  title={Identifying and quantifying the abundance of economically important palms in tropical moist forest using UAV imagery},
  author={Tagle Casapia, Ximena and Falen, Lourdes and Bartholomeus, Harm and C{\'a}rdenas, Rodolfo and Flores, Gerardo and Herold, Martin and Honorio Coronado, Eur{\'\i}dice N and Baker, Timothy R},
  journal={Remote Sensing},
  volume={12},
  number={1},
  pages={9},
  year={2019},
  publisher={MDPI}
}

@article{van2019palm,
  title={The palm Mauritia flexuosa, a keystone plant resource on multiple fronts},
  author={van der Hoek, Yntze and {\'A}lvarez Solas, Sara and Pe{\~n}uela, Mar{\'\i}a Cristina},
  journal={Biodiversity and Conservation},
  volume={28},
  number={3},
  pages={539--551},
  year={2019},
  publisher={Springer}
}

@incollection{terborgh1986community,
  title={Community aspects of frugivory in tropical forests},
  author={Terborgh, John},
  booktitle={Frugivores and seed dispersal},
  pages={371--384},
  year={1986},
  publisher={Springer}
}

@article{dosovitskiy2020image,
  title={An image is worth 16x16 words: Transformers for image recognition at scale},
  author={Dosovitskiy, Alexey and Beyer, Lucas and Kolesnikov, Alexander and Weissenborn, Dirk and Zhai, Xiaohua and Unterthiner, Thomas and others},
  journal={arXiv preprint arXiv:2010.11929},
  year={2020}
}

@inproceedings{selvaraju2017grad,
  title={Grad-cam: Visual explanations from deep networks via gradient-based localization},
  author={Selvaraju, Ramprasaath R and Cogswell, Michael and Das, Abhishek and Vedantam, Ramakrishna and Parikh, Devi and Batra, Dhruv},
  booktitle={Proceedings of the IEEE international conference on computer vision},
  pages={618--626},
  year={2017}
}

@article{browne2016diversity,
  title={Diversity of Palm Communities at Different Spatial Scales in a Recently Fragmented Tropical Landscape},
  author={Browne, Luke and Karubian, Jordan},
  journal={Botanical Journal of the Linnean Society},
  volume={182},
  number={2},
  pages={451--464},
  year={2016},
  publisher={Oxford University Press}
}

@article{lueder2022functional,
  title={Functional Traits, Species Diversity, and Species Composition of a Neotropical Palm Community Vary in Relation to Forest Age},
  author={Lueder, Sarah and Narasimhan, Kaushik and Olivo, Jorge and Cabrera, Domingo and Jurado, Juana and Greenstein, Lewis and Karubian, Jordan},
  journal={Frontiers in Ecology and Evolution},
  volume={10},
  pages={678125},
  year={2022},
  publisher={Frontiers}
}

@article{tao2021mapping,
  title={Mapping tropical forest trees across large areas with lightweight cost-effective terrestrial laser scanning},
  author={Tao, Shengli and Labri{\`e}re, Nicolas and Calders, Kim and Fischer, Fabian J{\"o}rg and Rau, E-Ping and Plaisance, Laetitia and Chave, J{\'e}r{\^o}me},
  journal={Annals of Forest Science},
  volume={78},
  number={4},
  pages={103},
  year={2021},
  publisher={Springer}
}

@article{ben2021spatial,
  title={Spatial point-pattern analysis as a powerful tool in identifying pattern-process relationships in plant ecology: an updated review},
  author={Ben-Said, Mariem},
  journal={Ecological Processes},
  volume={10},
  pages={1--23},
  year={2021},
  publisher={Springer}
}

@article{ravi2024sam,
  title={Sam 2: Segment anything in images and videos},
  author={Ravi, Nikhila and Gabeur, Valentin and Hu, Yuan-Ting and Hu, Ronghang and Ryali, Chaitanya and Ma, Tengyu and Khedr, Haitham and R{\"a}dle, Roman and Rolland, Chloe and Gustafson, Laura and others},
  journal={arXiv preprint arXiv:2408.00714},
  year={2024}
}

@article{cui2024superpixel,
  title={Superpixel-based and Spatially-regularized Diffusion Learning for Unsupervised Hyperspectral Image Clustering},
  author={Cui, Kangning and Li, Ruoning and Polk, Sam L and Lin, Yinyi and Zhang, Hongsheng and Murphy, James M and Plemmons, Robert J and Chan, Raymond H},
  journal={IEEE Transactions on Geoscience and Remote Sensing},
  year={2024},
  publisher={IEEE}
}

@inproceedings{he2022masked,
  title={Masked autoencoders are scalable vision learners},
  author={He, Kaiming and Chen, Xinlei and Xie, Saining and Li, Yanghao and Doll{\'a}r, Piotr and Girshick, Ross},
  booktitle={Proceedings of the IEEE/CVF conference on computer vision and pattern recognition},
  pages={16000--16009},
  year={2022}
}

@inproceedings{ryali2023hiera,
  title={Hiera: A hierarchical vision transformer without the bells-and-whistles},
  author={Ryali, Chaitanya and Hu, Yuan-Ting and Bolya, Daniel and Wei, Chen and Fan, Haoqi and Huang, Po-Yao and Aggarwal, Vaibhav and Chowdhury, Arkabandhu and Poursaeed, Omid and Hoffman, Judy and others},
  booktitle={International Conference on Machine Learning},
  pages={29441--29454},
  year={2023},
  organization={PMLR}
}

@article{hinton2015distilling,
  title={Distilling the knowledge in a neural network},
  author={Hinton, Geoffrey and Vinyals, Oriol and Dean, Jeff},
  journal={arXiv preprint arXiv:1503.02531},
  year={2015}
}

@article{di2019wilderness,
  title={Wilderness areas halve the extinction risk of terrestrial biodiversity},
  author={Di Marco, Moreno and Ferrier, Simon and Harwood, Tom D and Hoskins, Andrew J and Watson, James EM},
  journal={Nature},
  volume={573},
  number={7775},
  pages={582--585},
  year={2019},
  publisher={Nature Publishing Group}
}

@inproceedings{camalan2022detecting,
  title={Detecting change due to alluvial gold mining in peruvian rainforest using recursive convolutional neural networks and contrastive learning},
  author={Camalan, Seda and Cui, Kangning and Pauca, Pa{\'u}l and Alqahtani, Sarra and Silman, Miles and Chan, Raymond and Dethier, Evan and Fernandez, Luis E and Lutz, David A and Plemmons, Robert},
  booktitle={AGU Fall Meeting Abstracts},
  volume={2022},
  pages={B52G--0905},
  year={2022}
}

@inproceedings{cui2022semi,
  title={Semi-supervised change detection of small water bodies using RGB and multispectral images in peruvian rainforests},
  author={Cui, Kangning and Camalan, Seda and Li, Ruoning and Pauca, Victor Paul and Alqahtani, Sarra and Plemmons, Robert and Silman, Miles and Dethier, Evan Nylen and Lutz, David and Chan, Raymond},
  booktitle={2022 12th Workshop on Hyperspectral Imaging and Signal Processing: Evolution in Remote Sensing (WHISPERS)},
  pages={1--5},
  year={2022},
  organization={IEEE}
}

@article{camalan2022change,
  title={Change detection of amazonian alluvial gold mining using deep learning and sentinel-2 imagery},
  author={Camalan, Seda and Cui, Kangning and Pauca, Victor Paul and Alqahtani, Sarra and Silman, Miles and Chan, Raymond and Plemmons, Robert Jame and Dethier, Evan Nylen and Fernandez, Luis E and Lutz, David A},
  journal={Remote Sensing},
  volume={14},
  number={7},
  pages={1746},
  year={2022},
  publisher={MDPI}
}

@article{Petritan2014forest,
    title = {Overstory succession in a mixed Quercus petraea–Fagus sylvatica old growth forest revealed through the spatial pattern of competition and mortality},
    author = {Petritan, Ion and Marzano, Raffaella and Petritan, Any Mary and Lingua, Emanuele},
    year = {2014},
    pages = {9--17},
    volume = {326},
    journal = {Forest Ecology and Management}
}

@article{Gadow2012structure,
author = {Gadow, Klaus and Zhang, Chunyu and Wehenkel, Christian and Pommerening, Arne and Corral-Rivas, Jose Javier and Korol, Mykola and Myklush, Stepan and Hui, Gangying and Kiviste, Andres and Zhao, Xiu},
year = {2012},
pages = {29--83},
title = {Forest Structure and Diversity},
volume = {23},
journal = {Continuous Cover Forestry, Book Series Managing Forest Ecosystems}
}

@article{May2015Moving,
author = {May, Felix and Huth, Andreas and Wiegand, Thorsten},
year = {2015},
title = {Moving beyond abundance distributions: Neutral theory and spatial patterns in a tropical forest},
volume = {282},
journal = {Proceedings. Biological sciences / The Royal Society}
}

@article{Jia2016MechanismUT,
  title={Mechanism Underlying the Spatial Pattern Formation of Dominant Tree Species in a Natural Secondary Forest},
  author={Guodong Jia and Xinxiao Yu and Dengxing Fan and Jianbo Jia},
  journal={PLoS ONE},
  year={2016},
  volume={11}
}

@book{Ripley1981, 
    place={New York}, 
    title={Spatial statistics}, 
    publisher={Wiley}, 
    author={Ripley, Brian D.}, 
    year={1981}
}

@article{Velázquez2016evaluation,
    author = {Velázquez, Eduardo and Martinez Cano, Isabel and Getzin, Stephan and Moloney, Kirk and Wiegand, Thorsten},
    year = {2016},
    title = {An evaluation of the state of spatial point pattern analysis in ecology},
    volume = {39},
    journal = {Ecography}
}

@book{Baddeley2016, 
    place={Boca Raton}, 
    title={Spatial point patterns: Methodology and applications with R}, 
    publisher={CRC Press, Taylor \& Francis Group},
    author={Baddeley, Adrian and Rubak, Ege and Turner, Rolf}, 
    year={2016}
}

@article{Wiegand2004RingsCA,
    title={Rings, circles, and null-models for point pattern analysis in ecology},
    author={Thorsten Wiegand and Kirk A. Moloney},
    journal={Oikos},
    year={2004},
    volume={104},
    pages={209--229}
}

@article{Renner2015HP,
    author = {Renner, Ian and Elith, Jane and Baddeley, Adrian and Fithian, William and Hastie, Trevor and Phillips, Steven and Popovic, Gordana and Warton, David},
    year = {2015},
    title = {Point process models for presence‐only analysis},
    volume = {6},
    journal = {Methods in Ecology and Evolution}
}

@article{Carrer2018HP,
    author = {Carrer, Marco and Castagneri, Daniele and Ionel, Popa and Pividori, Mario and Lingua, Emanuele},
    year = {2018},
    pages = {125--134},
    title = {Tree spatial patterns and stand attributes in temperate forests: The importance of plot size, sampling design, and null model},
    volume = {417},
    journal = {Forest Ecology and Management}
}

@article{Law2009thomas,
    author = {Law, Richard and Illian, Janine and Burslem, David and Gratzer, Georg and Gunatilleke, C. and Gunatilleke, Nimal},
    year = {2009},
    pages = {616--628},
    title = {Ecoogical information from satial patterns of plants: Insights from point process theory},
    volume = {97},
    journal = {Journal of Ecology}
}

@article{Jácome2016thomas,
    author = {Jácome-Flores, Miguel and Delibes, Miguel and Wiegand, Thorsten and Fedriani, Jose},
    year = {2016},
    month = {11},
    pages = {1--13},
    title = {Spatial patterns of an endemic Mediterranean palm recolonizing old fields},
    volume = {6},
    journal = {Ecology and Evolution},
    doi = {10.1002/ece3.2504}
}

@book{kingman1992poisson,
  title={Poisson processes},
  author={Kingman, John Frank Charles},
  volume={3},
  year={1992},
  publisher={Clarendon Press}
}

@article{park2019quantifying,
  title={Quantifying leaf phenology of individual trees and species in a tropical forest using unmanned aerial vehicle (UAV) images},
  author={Park, John Y and Muller-Landau, Helene C and Lichstein, Jeremy W and Rifai, Sami W and Dandois, Jonathan P and Bohlman, Stephanie A},
  journal={Remote Sensing},
  volume={11},
  number={13},
  pages={1534},
  year={2019},
  publisher={MDPI}
}

@article{clark1999seed,
  title={Seed dispersal near and far: patterns across temperate and tropical forests},
  author={Clark, James S and Silman, Miles and Kern, Ruth and Macklin, Eric and HilleRisLambers, Janneke},
  journal={Ecology},
  volume={80},
  number={5},
  pages={1475--1494},
  year={1999},
  publisher={Wiley Online Library}
}

@article{metz2010widespread,
  title={Widespread density-dependent seedling mortality promotes species coexistence in a highly diverse Amazonian rain forest},
  author={Metz, Margaret R and Sousa, Wayne P and Valencia, Renato},
  journal={Ecology},
  volume={91},
  number={12},
  pages={3675--3685},
  year={2010},
  publisher={Wiley Online Library}
}

@article{K_L_func1,
    title = {Effects of sampling scale on patterns of habitat association in tropical trees},
    author = {Garzón López, Carol Ximena and Jansen, Patrick and Bohlman, Stephanie and Ordonez, Alejandro and Olff, Han},
    journal = {Journal of Vegetation Science},
    year = {2013},
    month = {05},
    volume = {25}
}

@article{Haase1995,
  title={Spatial pattern analysis in ecology based on Ripley's K-function: Introduction and methods of edge correction},
  author={Peter Haase},
  journal={Journal of Vegetation Science},
  year={1995},
  volume={6},
  pages={575--582}
}

@article{Goreaud2000,
    author = {François Goreaud},
    year = {2000},
    title = {Apports de l'analyse de la structure spatiale en forêt tempérée à l'étude et la modélisation des peuplements complexes},
    journal = {Dissertation},
    publisher={Centre de Nancy}
}

@article{Besag1977,
    author = {Julian Besag},
    title = {Contribution to the Discussion on Dr. Ripley’s Paper},
    journal = {Journal of the Royal Statistical Society: Series B (Methodological)},
    volume = {39},
    number = {2},
    pages = {193--195},
    year = {1977}
}

@book{hildebrand1987introduction,
  title={Introduction to numerical analysis},
  author={Hildebrand, Francis Begnaud},
  year={1987},
  publisher={Courier Corporation}
}

\end{document}